\documentclass[preprint,12pt]{elsarticle}

\usepackage{amssymb}

\journal{SIAM Journal on Imaging Sciences}
\usepackage{latexsym,bm,amsmath,amssymb}
\usepackage{enumerate}
\usepackage{graphicx}
\usepackage[ruled,vlined]{algorithm2e}
\usepackage{algorithmic}
\usepackage{amsfonts,amssymb,amsmath}
\usepackage{dsfont}
\usepackage{multirow}
\usepackage[table]{xcolor}
\usepackage{colortbl}
\usepackage[T1]{fontenc}
\usepackage{bm}
\usepackage{setspace}
\usepackage{caption}
\usepackage{color}

\usepackage{graphicx}
\usepackage{setspace}
\usepackage{float}
\usepackage{epsfig,epstopdf}
\usepackage{verbatim}
\usepackage{algorithmic}
\usepackage{array}
\usepackage{amsthm}
\usepackage{booktabs}
\usepackage{graphicx}
\usepackage{subfigure}
\usepackage{multirow}
\usepackage{threeparttable}
\usepackage{hyperref}
\usepackage{makecell}
\usepackage{fmtcount}
\usepackage{booktabs}

\usepackage{nicefrac}
\usepackage{xcolor}
\usepackage{xspace}
\usepackage{enumerate}
\usepackage{mathrsfs}



\newcolumntype{L}[1]{>{\raggedright\let\newline\\\arraybackslash\hspace{0pt}}m{#1}}
\newcolumntype{C}[1]{>{\centering\let\newline\\\arraybackslash\hspace{0pt}}m{#1}}
\newcolumntype{R}[1]{>{\raggedleft\let\newline\\\arraybackslash\hspace{0pt}}m{#1}}

\numberwithin{equation}{section}

\theoremstyle{remark}

\newcommand{\mat}[1]{\ensuremath{\mathbf{#1}}}

\newcommand{\cG}{\mathcal{G}}
\newcommand{\ident}{\mat{I}}

\newcommand{\eg}{e.g.}
\newcommand{\ie}{i.e.}
\newcommand{\etal}{\textit{et al.}}

\newcommand{\cL}{\mathcal{L}}
\newcommand{\cD}{\mathcal{D}}

\newcommand{\R}{\mathbb{R}}

\newcommand{\suml}[2]{\sum\nolimits_{#1}^{#2}}

\graphicspath{{figures/}}

\begin{document}
	
\begin{frontmatter}
\title{Image Denoising via Multi-scale Nonlinear Diffusion Models}


\author[labelFWS]{Wensen Feng}
\author[labelQP]{Peng Qiao}
\author[labelXY]{Xuanyang Xi}
\author[labelYJ]{Yunjin Chen\corref{cor3}}
\ead{chenyunjin$\_$nudt@hotmail.com}

\address[labelFWS]{School of Automation and Electrical Engineering, University of Science and Technology Beijing, Beijing 100083, China.}
\address[labelQP]{National Laboratory for Parallel
and Distributed Processing, School of Computer, National University of
Defense Technology, Changsha, 410073, China.}
\address[labelXY]{State Key Laboratory of Management and Control for Complex Systems, Institute of Automation, Chinese Academy of Sciences, Beijing 100190, China.}
\address[labelYJ]{Institute for Computer Graphics and Vision,
Graz University of Technology, 8010 Graz, Austria}
\cortext[cor3]{Corresponding author.}

\begin{abstract}
Image denoising is a fundamental operation in image processing and holds considerable
practical importance for various real-world applications. Arguably several
thousands of papers are dedicated to image denoising.
In the past decade, sate-of-the-art denoising algorithm have been clearly dominated
by non-local patch-based methods, which explicitly exploit patch self-similarity
within image. However, in recent two years, discriminatively trained local approaches
have started to outperform previous non-local models and have been attracting increasing
attentions due to the additional advantage of computational efficiency.
Successful approaches include cascade of shrinkage fields (CSF) and trainable
nonlinear reaction diffusion (TNRD). These two methods are built on filter response
of linear filters of small size using feed forward architectures.
Due to the locality inherent in local approaches,
the CSF and TNRD model become less effective when noise level is high and consequently
introduces some noise artifacts.
In order to overcome this problem, in this paper we introduce a multi-scale strategy.
To be specific, we build on our newly-developed TNRD model, adopting the multi-scale
pyramid image representation to devise a multi-scale nonlinear diffusion process.
As expected, all the parameters in the proposed multi-scale diffusion model,
including the filters and the influence functions across scales,
are learned from training data through a loss based approach.
Numerical results on Gaussian and Poisson denoising substantiate that
the exploited multi-scale strategy can successfully boost the performance of the
original TNRD model with single scale. As a consequence,
the resulting multi-scale diffusion models can
significantly suppress the typical incorrect features for those noisy images
with heavy noise. It turns out that multi-scale TNRD variants
achieve better performance than state-of-the-art denoising methods.
\end{abstract}


\end{frontmatter}
\section{Introduction}
Image denosing is a widely studied problem with immediate practical applications
in image processing and computer vision, serving as a preprocessing step.
During the past decades, many remarkable denoising methods have been proposed
to improve the performance of single-image based denoising.
Up to now, image denoising still remains a vibrant research field
\cite{chatterjee2010denoising}. Roughly speaking, image denoising techniques
can be classified into two categories: (1) nonlocal methods, which exploit nonlocal
similarity information within long range distance,
such as \cite{dabov2007image, gu2014weighted},
or global information in the whole image \cite{talebi2014global} or even similar patches from
an external dataset \cite{yue2015image}; and (2) local methods, which solely depend on
information in a local neighborhood, such as \cite{roth2009fields, elad2006image}.

For the present, most existing state-of-the-art image denoising algorithms are based
on exploiting nonlocal similarity between a relatively modest number of patches,
for instance, Gaussian denoising \cite{mairal2009non, gu2014weighted, dabov2007image},
multiplicative noise reduction \cite{cozzolino2014fast}, \cite{parrilli2012nonlocal} and
Poisson noise suppression \cite{salmon2014poisson}, \cite{giryes2014sparsity}.
The basic idea of the nonlocal approaches is to estimate the target pixel based on a
carefully selected set of pixels: not just those closest to the target but also
those distant pixels that are more likely to have the same underlying signal.
Usually, the goal of nonlocal techniques mainly concentrate on achieving
utmost image restoration quality, but with little consideration
on the computational efficiency. As a consequence, predominant state-of-the-art
image denoising approaches are without a doubt non-local models.

In contrast to nonlocal denoising approaches, local methods merely investigate
the local information from the surrounding pixels to denoise the central pixel.
Since the well-known P-M model \cite{perona1990scale} and ROF model
\cite{rudin1992nonlinear}, many attempts have been made to develop effective local
models, such as Product of Experts (PoT) \cite{welling2002learning},
Fields of Experts (FoE) framework \cite{roth2009fields} and K-SVD \cite{elad2006image}.
Due to the inherent drawback of locality, one would not expect
a local model can compete with those good-performing nonlocal approaches. However, in
recent years a few local models, \eg, \cite{zoran2011learning, burger2012image,
chen2014insights}, obtained from elaborate learning are able to produce
comparable results to state-of-the-arts, such as BM3D \cite{dabov2007image}.
Especially, two newly-developed local models - the CSF \cite{schmidt2014shrinkage} and
TNRD model \cite{chenCVPR15}, which are also discriminatively trained,
successfully achieve superior denoising performance to previous state-of-the-arts.
Therefore, discriminative model learning for image denoising has been attracting
more and more attentions as it additionally bears a significant merit
of computational efficiency.

\begin{figure}[t!]
\centering
\subfigure[Clean Image]{
\includegraphics[width=0.2\textwidth]{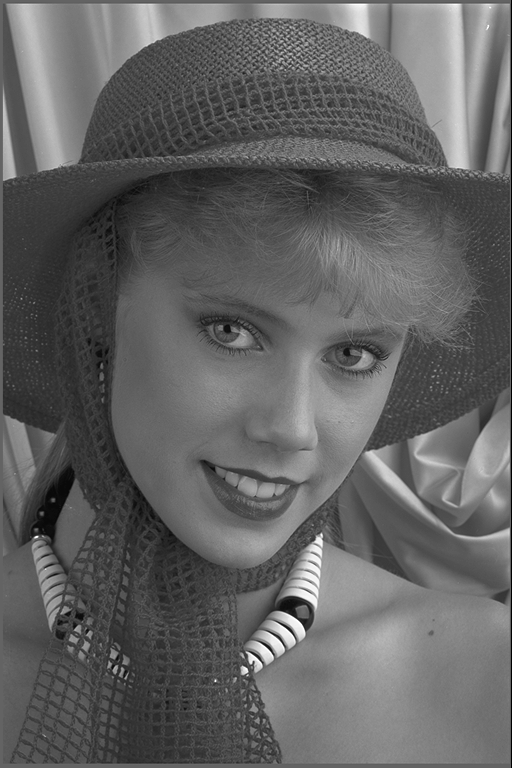}
}
\subfigure[Noisy image. $\sigma=50$]{
\includegraphics[width=0.2\textwidth]{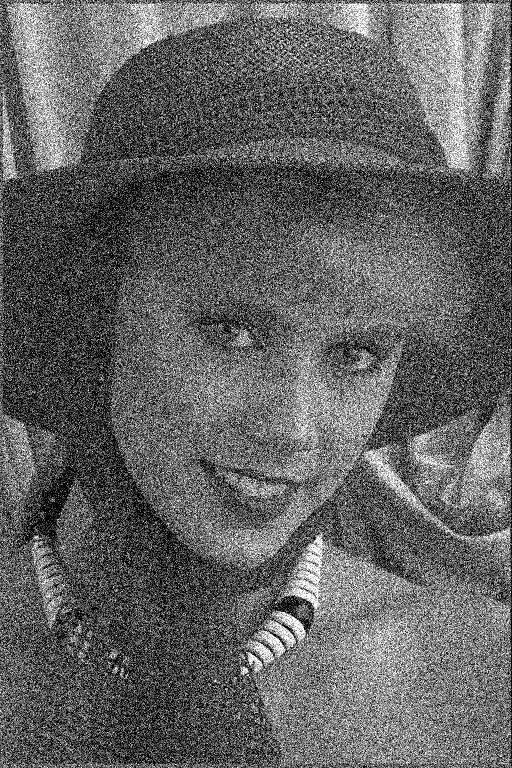}
}
\subfigure[$\mathrm{TNRD}_{5 \times 5}^5$ (28.58/0.718)]{
\includegraphics[width=0.2\textwidth]{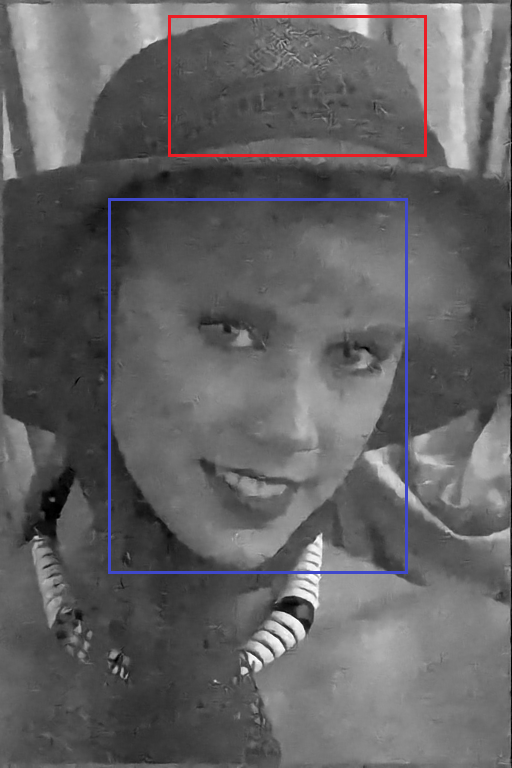}
}
\subfigure[$\mathrm{MSND}_{5 \times 5}^5$ (\textbf{28.95}/\textbf{0.730})]{
\includegraphics[width=0.2\textwidth]{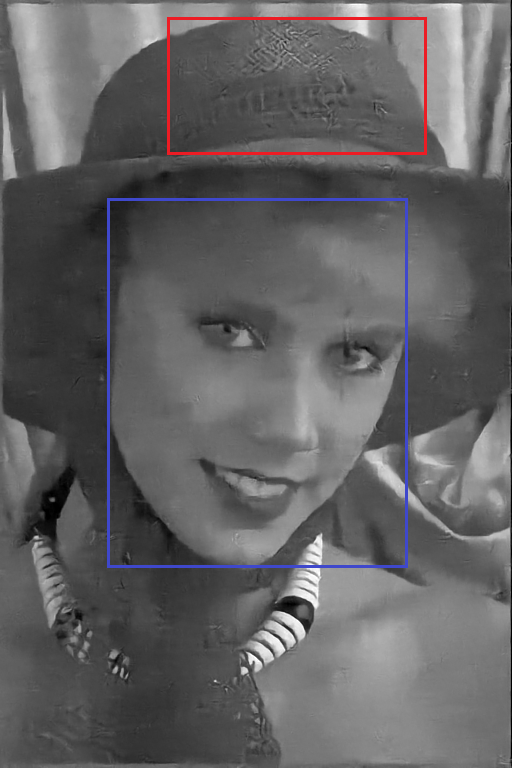}
}
\caption{Gaussian noise denoising results comparison. The results are reported by PSNR/SSIM index. Better result is marked. We can see that MSND model achieves apparent improvements in the highlighted regions, such as face and hat.}
\label{comparetest}
\end{figure}

As aforementioned, local models infer the underlying structure solely from the local
neighborhoods. Despite the general effectiveness of the CSF and TNRD model, they can not
avoid this inherent drawback. As a consequence, when the noise level is relatively high,
some incorrect image features, which are not in the clean image, appear in the denoised
results. In this case, local image structures are heavily distorted by noise (as shown in Fig.~\ref{comparetest})(c), and
therefore it becomes very challenging to estimate these structures.
Under this circumstance, a larger region containing larger-scale information
would benefit for effective denoising.

A deeper inspection of the CSF and TNRD model reveals that it is the small size of
involved filters that causes these approaches to ignore larger-scale information.
So we hypothesize that these two denoising algorithms can be improved by employing
a multi-scale strategy.

\subsection{Multi-scale strategy}
Multi-scale analysis, which becomes a standard tool in image analysis,
has been widely exploited for diverse image processing problems,
such as image segmentation \cite{liu2014mslrr}, hyper-spectral image classification
\cite{kang2014spectral}, single image defogging \cite{wang2014single}, single image
dehazing \cite{ancuti2013single} and the image denoising task investigated in this paper
\cite{lebrun2015multiscale, zontak2013separating, burger2011improving}. Generally speaking,
multi-scale analysis can represent distinct but complementary information
that exists at coarse-to-fine resolution levels. Therefore, such representations
reveal more structural information about signals and offer a number of advantages
over fixed-scale methods.

In the context of Wavelet transform based image processing,
multi-scale analysis can be naturally incorporated
as the Wavelet transforms usually decompose an image in a multi-scale
manner. A few Wavelet based image denoising algorithms can deliver appealing
performance \cite{chang2000adaptive, portilla2003image} via the well-known Wavelet
shrinkage operation.

In the case of image denoising,
multi-scale analysis is more commonly understood as image pyramid, \ie, down-sampling the
input image to build the noisy pyramid. As pointed out in
\cite{lebrun2015multiscale, burger2011improving} that multi-scale analysis
would be generally beneficial to image denoising, as it promises at least two improvements
\begin{itemize}
    \item[1)] Down-sampling has a denoising effect. For the particular case of
Gaussian white noise, consider down-sampling a noisy image
by a factor of two in the way that distinct blocks of four pixels are averaged to form
a single pixel in the low-resolution image, the standard deviation of the noise
will be divided by two compared to the noise contained in original noisy image.
Therefore, down-sampling can cause the uncorrelated values of the noise to
become smaller and consequently make image structures more visible, which can be clearly observed by visual comparison on (c) and (d) in Fig.~\ref{comparetest}.
    \item[2)] Down-sampling the image before denoising amounts to enlarge the size of
the neighborhood on which the denoising is performed, thus permitting to
exploit larger-scale information. This will be particularly helpful for those
local models considered in this paper.
\end{itemize}

Due to the above-mentioned potential benefits from the multi-scale strategy, it has
been employed by many image denoising algorithms. Zontak \etal proposed using
the patch recurrence across scales to separate signal from noise
\cite{zontak2013separating}. In \cite{lebrun2015multiscale}, a multi-scale scheme
was exploited for blind denoising, and a multi-scale meta-procedure was considered to
improve the performance of a few denoising algorithms \cite{burger2011improving}.
In \cite{liu2014progressive} and \cite{lefkimmiatis2009bayesian},
a multi-scale representation was constructed to perform progressive image denoising and
Poisson denoising, respectively.

\subsection{Motivation and contributions}
The above-mentioned works substantiate the usefulness of the multi-scale structure.
Therefore, it is intriguing to incorporate the multi-scale strategy in the newly
developed TNRD framework \cite{chenCVPR15}
to investigate whether it can also boost the performance of the TNRD model with single scale.
In this paper, we only concentrate on the TNRD framework, as it exhibits a remarkable
advantage of high efficiency over existing denoising algorithms,
besides its compelling denoising performance.

In this work, we incorporate the multi-scale strategy into the TNRD framework, resulting in
a novel image denoising approach named as Multi-scale Nonlinear Diffusion Model (MSND).
Thus, the image diffusion process happens in a few scales instead of in a single scale,
and the resulting images are combined into a single denoised image in the finest scale via
up-sampling. As usual in discriminative model learning for image denoising,
all the parameters in the proposed multi-scale model, including the filters and the
influence functions, are supervised learned end-to-end with training samples.

In order to verify the validity of the multi-scale strategy in the TNRD framework for
image denoising, we investigate denoising problems for two typical types of noise:
Gaussian noise and Poisson noise.
Numerical experiments on Gaussian and Poisson denoising demonstrate that the resulting
multi-scale TNRD models can considerably improve the performance of the
single scale counterparts, especially when the noise level is high, therein the
multi-scale TNRD models can significantly suppress the typical incorrect features
appearing in the original TNRD models, as shown within the blue rectangle in Fig.~\ref{comparetest}(c) and (d).
It turns out that the resulting MSND models yield better performance than
state-of-the-art denoising approaches.

To summarize, the main contributions of this paper are three folds.
\begin{itemize}
    \item[1)] We extend the TNRD framework with a multi-scale strategy, which can
significantly enlarge the size of local receptive field exploited for diffusion;
    \item[2)] We demonstrate the versatility of the TNRD framework and its multi-scale
variant through an additional image denoising problem at the presence of Poisson noise;
    \item[3)] Extensive experimental results substantiate the usefulness of the
multi-scale procedure, especially in the case of heavy noise. It turns out that the
multi-scale nonlinear diffusion models can surpass state-of-the-art denoising algorithms.
\end{itemize}

The remainder of the paper is organized as follows. Section II presents a general review of the trainable nonlinear reaction diffusion process. In the subsequent section III, we propose the Image Denoising approach via Multi-scale Nonlinear Diffusion, and take it for Gaussian noise removing and Poisson noise removing. Subsequently, Section IV describes comprehensive experiment results for the proposed model.
The concluding remarks are drawn in the final Section V.

\section{Preliminaries}
To make the paper self-contained, in this section we provide a brief review of
the trainable nonlinear diffusion
process proposed in \cite{chenCVPR15}.
\subsection{Highly parametrized nonlinear diffusion model}
In our recent work \cite{chenCVPR15}, a simple but effective framework for
image restoration called TNRD was proposed based on the concept of
nonlinear reaction diffusion.
The TNRD framework is modeled by highly parametrized
linear filters as well as highly parametrized
influence functions. In contrast to those conventional nonlinear diffusion models which
usually make use of handcrafted parameters, all the parameters in the TNRD model,
including the filters and the influence functions, are learned from training
data through a loss based approach.

The general framework of the TNRD model
is formulated as the following time-dynamic nonlinear reaction-diffusion process
with $T$ steps
\begin{equation}\label{diffusion}
\small
\hspace{-0.25cm}
\begin{cases}
u_0 = I_0 \\
u_{t} = \underbrace{\text{Prox}_{\cG^t}}_\text{reaction force}\left(
u_{t-1} - \left(\underbrace{\sum\limits_{i = 1}^{N_k}
(K_i^t)^\top \phi_i^t(K_i^t u_{t-1})}_\text{diffusion force} + \underbrace{\psi^t(u_{t-1}, f)}_
\text{reaction force}\right)\right)\,, \\
t = 1 \cdots T \,,
\end{cases}
\end{equation}
where image $u$ is represented as a column vector, \ie, $u \in \R^{N}$,
$I_0$ is the initial status of the diffusion process and
$T$ denotes the diffusion stages.
$K_i \in \R^{N \times N}$ is a highly sparse matrix, implemented as
2D convolution of the image $u$
with the linear filter $k_i$, i.e., $K_i u \Leftrightarrow k_i*u$,
$K_i$ is a set of linear filters and $N_k$ is the number of filters.
In this formulation, function $\phi_i$ is known as influence function and
is applied point-wise to the filter response, \ie,
$\phi(K_i u)= \left( \phi(K_i u)_1, \cdots, \phi(K_i u)_N \right)^\top
\in \R^N$.

Both the proximal mapping operation $\text{Prox}_{\cG}(\hat u)$
and $\psi(u_t, f)$ are related to the reaction force. The proposed TNRD
can be applied to various image restoration problems by exploiting
task-specific function $\cG(u, f)$ and $\psi(u, f)$.
Usually, the reaction term $\psi(u)$ is designated as the derivative of a certain
smooth date term $\cD(u, f)$, i.e., $\psi(u) = \nabla_u \cD(u)$. The
term $\cG(u, f)$ is used to handle those problems with a non-smooth date term,
\eg, JPEG deblocking in \cite{chenCVPR15} or a data term for which a
gradient descent step is not appropriate, \eg, Poisson denoising investigated
in this paper. Note that
the proximal mapping operation \cite{nesterov2004introductory} related to
the function $\cG^t$ is given as
\[
\text{Prox}_{\cG^t}(\hat u) = \min\limits_{u}\frac{\|u - \hat u\|_2^2}{2}
+ \cG^t(u, f) \,.
\]

The diffusion term involves trainable linear filters and influence functions
(nonlinearities) and
these parameters vary in each iteration, i.e., time varying linear filters
and nonlinearities.

As shown in \cite{chenCVPR15},
the proposed model \eqref{diffusion} can be interpreted as
performing one gradient descent step at $u^t$ with respect to a dynamic energy
functional given by
\begin{equation}\label{foemodel}
E^t(u, f) = \suml{i = 1}{N_k}\sum\limits_{p = 1}^{N} \rho_i^t((k_i^t
*u)_p) + \cD^t(u, f) + \cG^t(u, f)\,,
\end{equation}
where the functions $\left\{\rho_i^t\right\}_{t=1}^{t=T}$ are the so-called penalty
functions. Note that $\rho'(z) = \phi(z)$ and the parameters \{$k_i^t,
\rho_i^t$\} vary across the stages i.e., changes at each iteration.

The TNRD model also bears an interesting link
to convolutional networks (CNs) applied to image restoration
problems in \cite{CNNdenoising}. For example, from the architecture of our
proposed diffusion model for Gaussian denoising,
shown in Figure \ref{fig:feedforwardCNN},
one can see that each iteration (stage) of our
proposed diffusion process involves convolution operations with a
set of linear filters, and thus it can be treated as a convolutional
network. The most noticeable aspect in the proposed convolutional
network is that the nonlinearities (i.e., influence functions
in the context of nonlinear diffusion) are trainable, instead of
a fixed activation function, e.g., the ReLU function \cite{nair2010rectified} or
sigmoid functions \cite{CNNdenoising}, in  conventional CNs.

The TNRD framework is applicable to different image restoration problems
by incorporating specific reaction force, such as
Gaussian denoising and Poisson denoising investigated in this paper.
For Gaussian denoising, it is realized by setting $\cG^t = 0$, while $\cD^t(u, f) =
\frac{\lambda^t}{2} \|u - f\|_2^2$ and $\psi^t(u) = \lambda^t (u -
f)$, where $\lambda^t$ is related to the strength of the reaction term,
$f$ denote the input degraded image.

In the case of Poisson denoising, taking into account the peculiar features of
Poisson noise, the relevant data term is given by
the so-called Csisz{\'a}r I-divergence model
\cite{csiszar1991least, Le1, giryes2014sparsity}, defined as
\begin{equation}
\langle u-f \mathrm{log}u,1 \rangle,
\label{poissonfidelity}
\end{equation}
where $\langle,\rangle$ denotes the standard inner product. Then, in the proposed
model, we set $\cD^t = 0$, while $\cG^t(u, f) =
{\lambda^t} \langle u-f \mathrm{log}u,1 \rangle$. More details can be found
in Sec. \ref{sec:poisson}.

\subsection{Overall training scheme}
The TNRD model in \eqref{diffusion} is trained in a supervised manner, namely,
the input/output pairs
for certain image processing task are firstly prepared, and then
we exploit a loss minimization scheme to
optimizet the model parameters $\Theta^t$ for each
stage $t$ of the diffusion process. The training dataset consists of
$S$ training samples $\{u_{gt}^s,f^s\}_{s=1}^S$, where
$u_{gt}^s$ is a ground truth image and $f^s$ is the corresponding degraded input.
The model parameters in stage $t$ required
to be trained include 1) the reaction force weight $\lambda$,
(2) linear filters and (3) influence functions. All parameters are grouped
as $\Theta^t$, i.e., $\Theta^t = \{\lambda^t, \phi_i^t, k_i^t\}$.
Then, the optimization problem for the training task is formulated as follows
\begin{equation}\label{learning}
\small
\hspace{-0.25cm}
\begin{cases}
\min\limits_{\Theta}\cL(\Theta) = \sum\limits_{s = 1}^{S} \ell(u_T^s, u_{gt}^s) =
\sum\limits_{s = 1}^{S}\frac 1 2
\|u_T^s - u_{gt}^s\|^2_2\\
\text{s.t.}
\begin{cases}
u_0^s = I_0^s \\
u_{t}^s = \text{Prox}_{\cG^t}\left(u_{t-1}^s - \left({\sum\limits_{i = 1}^{N_k}
(K_i^t)^\top \phi_i^t(K_i^t u_{t-1}^s)} + \psi^t(u_{t-1}^s, f^s)
\right)\right), \\
t = 1 \cdots T\,,
\end{cases}
\end{cases}
\end{equation}
where $\Theta = \{\Theta^t\}_{t=1}^{t=T}$.
The training problem in \eqref{learning} can be solved via gradient
based algorithms, e.g., the L-BFGS algorithm \cite{lbfgs}, where the gradients associated with $\Theta_t$ are computed
using the standard back-propagation technique \cite{lecun1998gradient}.

There are two training strategies to learn the diffusion processes:
1) the greedy training strategy to learn the diffusion process stage-by-stage;
and 2) the joint training strategy to train a diffusion process by simultaneously tuning
the parameters in all stages. Generally speaking, the joint training strategy
performs better \cite{chenCVPR15}, and the greedy training strategy is often
used to provide a good initialization for the joint training.
Concerning the joint training scheme, the gradients $\frac {\partial
\ell(u_T, u_{gt})}{\partial \Theta_t}$ are computed as follows,
\begin{equation}
\frac {\partial \ell(u_T, u_{gt})}{\partial \Theta_t} =
\frac {\partial u_t}{\partial \Theta_t} \cdot \frac {\partial u_{t+1}}{\partial u_{t}} \cdots
\frac {\partial \ell(u_T, u_{gt})}{\partial u_T} \,,
\label{iterstep}
\end{equation}
which is known as back-propagation.
For different image restoration problems, we mainly need to
recompute the two components $\frac {\partial u_t}{\partial \Theta_t}$ and
$\frac {\partial u_{t+1}}{\partial u_{t}}$, the main part of which
are similar to the derivations in \cite{chenCVPR15}.

\section{The Proposed Multi-scale Nonlinear Diffusion Model}
\begin{figure}[t!]
\vspace*{-2cm}
\centering
{\includegraphics[width=0.8\linewidth]{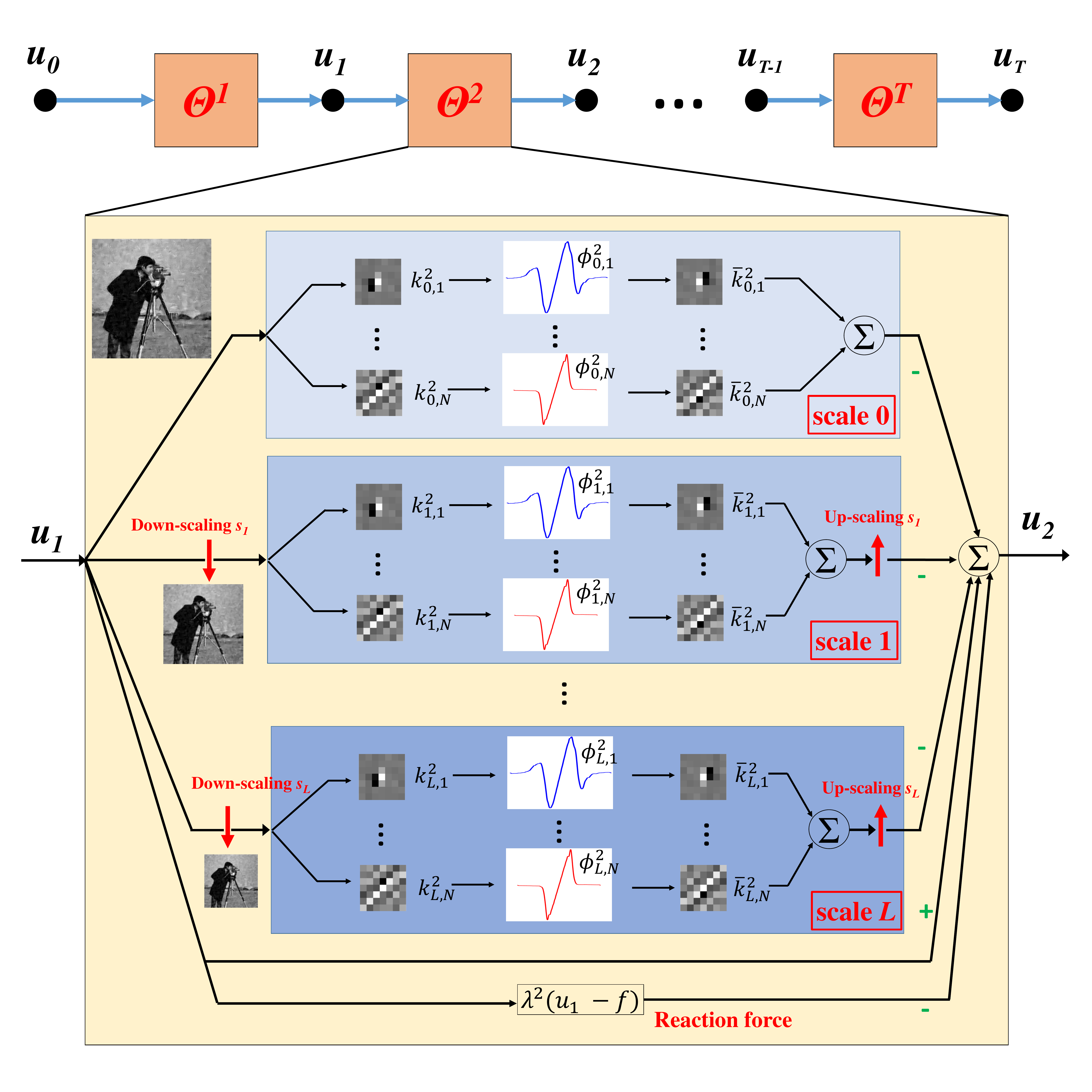}}
\caption{The architecture of the proposed multi-scale diffusion model for Gaussian
denoising. It is represented as a multi-layer feed-forward network.
Note that the additional convolution step with the rotated kernels $\bar k_i$
(\textit{cf.} Equ. \ref{Mg-diffusion}) does not appear in conventional
feed-forward CNs. }\label{fig:feedforwardCNN}
\end{figure}
When the noise level is relatively high, the TNRD model will introduce some
artifacts in the denoised images. The resulting artifacts are more apparent
in the homogeneous regions of the image, typical of commonplace images
such as scenery and landscapes. As aforementioned, in order to alleviate
this phenomenon, in this paper, we introduce a multi-scale version of the TNRD algorithm,
namely, we adopt the multi-scale
pyramid image representation to devise a multi-scale nonlinear diffusion process.

\subsection{General formulation of the proposed multi-scale diffusion model}
The basic idea is to extend the diffusion term in the TNRD model \eqref{diffusion} to
a multi-scale version. As a result, the new diffusion term in the stage $t$ is given as
\begin{equation}\label{mdiffusion}
\sum\limits_{l = 0}^{L} \sum\limits_{i = 1}^{N_k}
A_l^\top (K_{l,i}^t)^\top \phi_{l,i}^t(K_{l,i}^t A_l u_{t-1})\,,
\end{equation}
where $L$ is the number of total scales exploited in the proposed model, and
the additional linear operator $A_l$ is related to down-sampling the original
image $u$ to different sizes, \ie, different scales. For example, $A_0$ is the identity
matrix without any change to the original image and
$A_l$ corresponds to down-sampling the original image with a factor
$s^{l-1}$, where $s$ is the scale factor for the adjacent image in the scale space.
In this paper, the down-sampling operator $A_l$ is constructed in the following form,
taking for instance the scale factor of 2
\[
\begin{array}{c}
A_l\left[\begin{array}{cccc}
u_{11} & u_{12} & u_{13} & u_{14} \\
u_{21} & u_{22} & u_{23} & u_{24} \\
u_{31} & u_{32} & u_{33} & u_{34} \\
u_{41} & u_{42} & u_{43} & u_{44}
\end{array}\right]
=
\frac{1}{4}\left[ \begin{array}{cc}
{u_{11}} + {u_{12}} + {u_{21}} + {u_{22}} & {u_{13}} + {u_{14}} + {u_{23}} + {u_{24}}\\
{u_{31}} + {u_{32}} + {u_{41}} + {u_{42}} & {u_{33}} + {u_{34}} + {u_{43}} + {u_{44}}
\end{array} \right].
\end{array}
\]

For other cases, the formulation of $A_l$ can be similarly derived.
Note that in \eqref{mdiffusion} the operator $A_l^\top$ refers to the corresponding
up-sampling operator to enlarge the low resolution image to the original size.

In summary, the proposed multi-scale framework is formulated as the time-dynamic nonlinear
reaction-diffusion process with $T$ steps shown in \eqref{multi-diffusion}.
In the training phase, this new diffusion process is plugged into the training
framework \eqref{learning} to replace the single scale based diffusion procedure.
The parameters to be trained in stage $t$ include
$\{\{\phi_{l,i}^t, k_{l,i}^t\}_{l=0}^{L} \}_{i=1}^{N_k}$
and the parameter $\lambda^t$ associated with the reaction force $\psi^t(u_{t-1}, f)$
or $\cG(u,f)$, which are grouped into to $\Theta^t = \left\{
\{\{\phi_{l,i}^t, k_{l,i}^t\}_{l=0}^{L} \}_{i=1}^{N_k}, \lambda^t\right\}$.
Remember that these parameters are optimized with a gradient-based algorithm, and
therefore, we have to compute the gradients of the loss function with respect to
$\Theta^t$. According to the basic chain rule \eqref{iterstep},
we mainly need to compute the derivations of
$\frac {\partial u_t}{\partial \Theta^t }$ and
$\frac {\partial u_{t+1}}{\partial u_{t}}$, as
the gradient $\frac {\partial \ell(u_T, u_{gt})}{\partial u_T}$ is simply given as
$
\frac {\partial \ell(u_T, u_{gt})}{\partial u_T} = u_T -  u_{gt}\,.
$
\begin{figure*}[t]
\begin{multline}
\label{multi-diffusion}
\begin{cases}
u_{t} = \underbrace{\text{Prox}_{\cG^t}}_\text{reaction force}\left(
u_{t-1} - \left(\underbrace{\sum\limits_{l = 0}^{L} \sum\limits_{i = 1}^{N_k}
A_l^\top (K_{l,i}^t)^\top \phi_{l,i}^t(K_{l,i}^t A_l u_{t-1})
}_\text{diffusion force} + \underbrace{\psi^t(u_{t-1}, f)}_
\text{reaction force}\right)
\right)\,, \\
\text{with}\quad u_0 = I_0, \quad t = 1 \cdots T \,.
\end{cases}
\end{multline}
\vspace{-1cm}
\end{figure*}

In the following subsections, we present the main results of the gradients for two
exploited denoising problems, and the detailed derivation steps are
referred to the notes for diffusion network \cite{chenCVPR15}.

\subsection{Gradients for the case of Gaussian denoising}
For the Gaussian denoising problem, the multi-scale diffusion process is given as
\begin{equation}\label{Mg-diffusion}
\small
u_{t} = u_{t-1} - \left(\sum\limits_{l = 0}^{L} \sum\limits_{i = 1}^{N_k}
A_l^\top \left(\bar k_{l,i}^t * \phi_{l,i}^t(k_{l,i}^t * A_l u_{t-1})\right)
 + \lambda^t (u_{t-1} - f)\right)\,,
\end{equation}
where the convolution kernel $\bar k_i$ (obtained by rotating the kernel
$k_i$ 180 degrees) is explicitly used to replace $K_i^\top$ for the
sake of model simplicity. In this formulation, image $u$ is understood as
a 2D matrix. The corresponding diffusion procedure is illustrated in
Figure \ref{fig:feedforwardCNN}.
Note that notation $A_l$ in this formulation is in a little misusage.
We should bear in mind that when operating with $A_l$ and $A_l^\top$,
the image $u$ is understood as a column vector.

In the following derivations, as we frequently encounter
the convolution operation $k *u$ (image
$u \in \R^{m \times n}$, $k \in \R^{r \times r}$), it is helpful to exploit the equivalence for convolution:
$
k*u \Longleftrightarrow Ku \Longleftrightarrow Uk \,.
$
That is to say, $k *u$ is equivalent to the matrix-vector product formulation $Ku$,
where $K \in \R^{N \times N}$ is a highly sparse matrix and $u$ is
a column vector $u \in \R^N$ with $N = m \times n$.
It can also be interpreted with $Uk$, where matrix $U \in \R^{N \times R}$ is
constructed from image $u$
and $k$ is a column vector $k \in \R^{R}$ with $R = r \times r$.
The latter form is particularly helpful for the computation
of the gradients of the loss function with respect to
the kernel $k$, as $U^\top v$ ($v \in \R^{N}$ is a column vector) can be explicitly
interpreted as a convolution operation,
which is widely used in classic convolutional neural networks \cite{bouvrie2006notes}.
Furthermore, we consider the {denominator layout notation} for matrix calculus
required for the computation of derivations, where the basic chain rule
is given as the following order
\[
\frac{\partial z}{\partial x} = \frac{\partial y}{\partial x} \cdot
\frac{\partial z}{\partial y}\,.
\]

\subsubsection{Computation of the derivatives of $\frac {\partial u_t}{\partial
k_{l,i}^t }$}

By analyzing the formulation \eqref{Mg-diffusion},
it is easy to see the following relationship
\[
u_t \to -A_l^\top \left(\underbrace{\bar k_{l,i}^t}_{h} *
\underbrace{\phi_{l,i}^t(k_{l,i}^t * A_l u_{t-1})}_{v}\right)
\]
where $h$ and $v$ are auxiliary variables defined as $h=\bar k_{l,i}^t$ and
$v = \phi_{l,i}^t(k_{l,i}^t * u_{t-1}^{(l)})$ with $u_{t-1}^{(l)} = A_l u_{t-1}$,
\ie, the down-sampled version of $u_{t-1}$. Then we can obtain
$\frac {\partial u_t}{\partial k_{l,i}^t}$ as follows
\begin{equation}\label{grad_uk}
\frac {\partial u_t}{\partial k_{l,i}^t} =
-\left(P_{inv}^\top V^\top + {U_{t-1}^{(l)}}^\top
\Lambda {\bar{K_{l,i}^t}}^\top\right)A_l\,,
\end{equation}
where $\Lambda$ is a diagonal matrix
$\Lambda = \text{diag}({\phi_{l,i}^t}'(z_1), \cdots, {\phi_{l,i}^t}'(z_p))$
(${\phi_{l,i}^t}'$ is the first order derivative of function ${\phi_{l,i}^t}$) with
$z = k_{l,i}^t * u_{t-1}^{(l)}$. In this formulation, we have employed
the equivalences
\[
h*v \Longleftrightarrow Hv \Longleftrightarrow Vh \,,
\, \mathrm{and}\,\,
z = k_{l,i}^t * u_{t-1}^{(l)} \Longleftrightarrow U_{t-1}^{(l)} k_{l,i}^t\,.
\]

In equation \eqref{grad_uk}, matrix $P_{inv}^\top$ is a linear operator which inverts
the vectorized kernel $k$. When the kernel $k$ is in the form of 2D matrix,
it is equivalent to the Matlab command
$
P_{inv}^\top k \Longleftrightarrow rot90(rot90(k)) \,.
$

In practice, we do not need to explicitly construct the matrices
$V, U_{t-1}^{(l)}, \bar{K_{l,i}^t}$.
Recall that the product of matrices
$V^\top, {U_{t-1}^{(l)}}^\top$ and a vector can be computed by the convolution
operator \cite{bouvrie2006notes}. As shown
in a previous work \cite{chen2014insights}, ${\bar{K_{l,i}^t}}^\top$
can be computed using the convolution operation
with the kernel $\bar k_{l,i}^t$ with careful boundary handling.

\begin{figure}[t!]
\vspace{-0.2cm}
\centering
\includegraphics[width=0.1\textwidth]{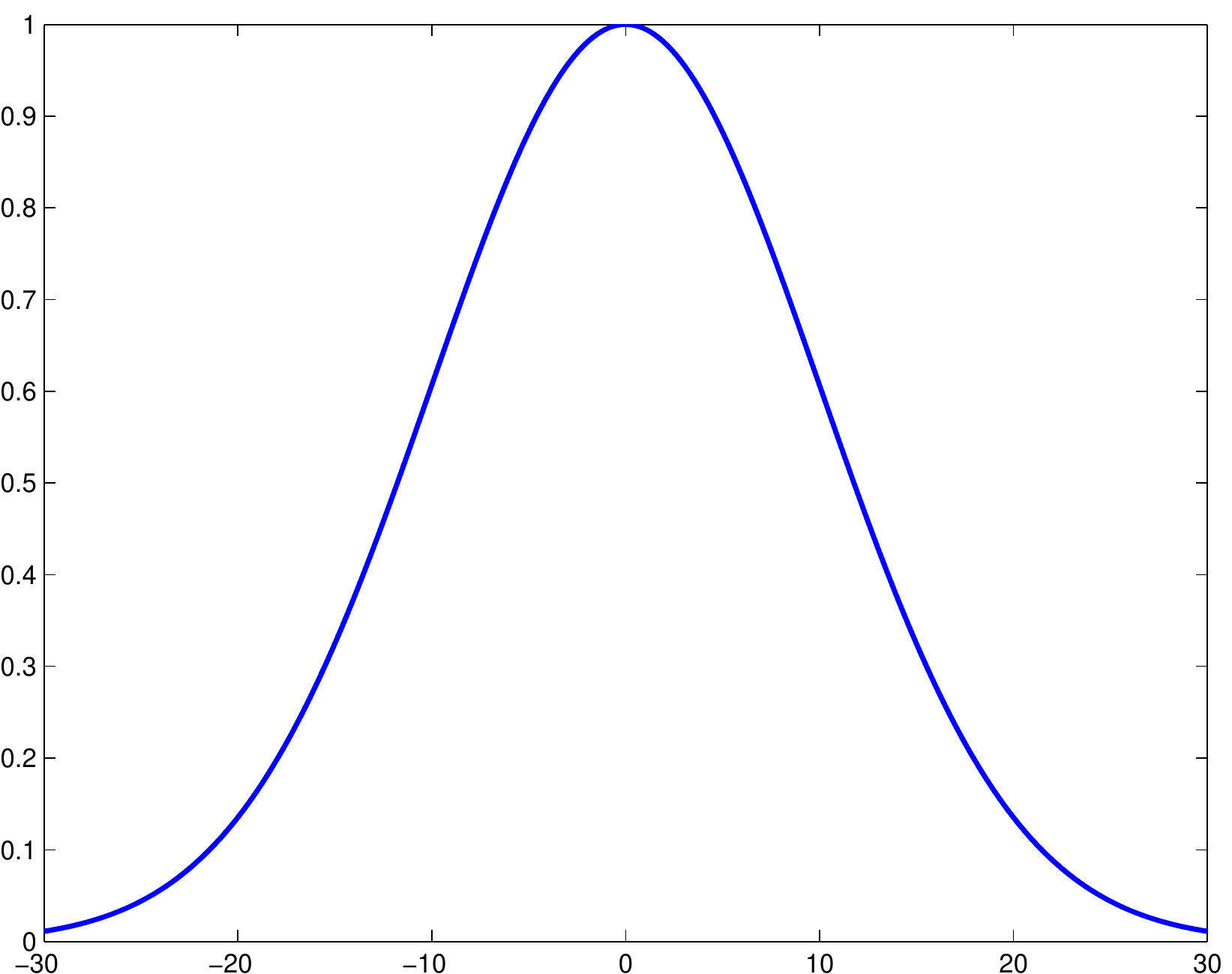}
\includegraphics[width=0.5\textwidth]{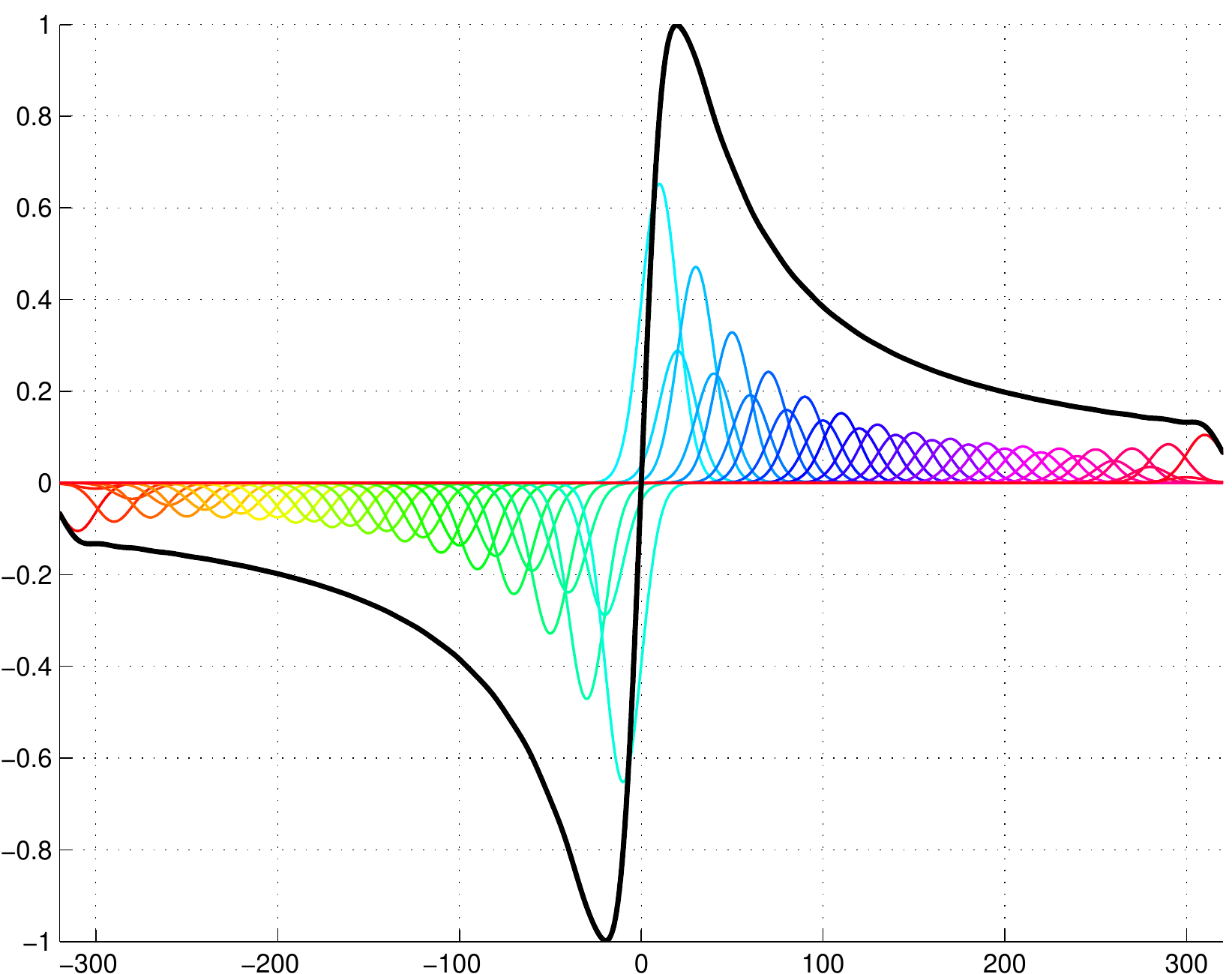}
\caption{Function approximation via Gaussian $\varphi_g(z)$
radial basis function.}\label{mapping}
\vspace{-0.4cm}
\end{figure}

\subsubsection{Computation of the derivatives of $\frac {\partial u_{t}}
{\partial \phi_{l,i}^t}$}
According to diffusion equation \eqref{Mg-diffusion}, the dependency of $u_t$ on the influence function $\phi_{l,i}^t$ is given as
\begin{equation}\label{xx}
u_t \to -{A_l}^{\top} \bar K_{l,i}^t \phi_{l,i}^t(z) \,,
\end{equation}
where $z = K_{l,i}^t A_l u_{t-1}$.
In this study, we consider the linear combination of
Gaussian Radial Basis Function (gRBF) for
function approximation. Thus, the function $\phi_{l,i}^t$ is represented as
\begin{equation}\label{rbf}
\phi_{l,i}^t(x) = \suml{j = 1}{M}w_{l,i,j}^t\varphi
\left(\frac {|x - \mu_j|}{\gamma}\right) \,,
\end{equation}
where $\varphi$ denotes the Gaussian kernel as shown in Fig.~\ref{mapping}, $M$ is the
number of basis kernels and $(\mu_j, \gamma)$ is the center of the $j^{th}$ basis and
the width, respectively.
Therefore, the vector $\phi_{l,i}^t(z)$ can be reformulated via a matrix equation
\[
\phi_{l,i}^t(z) = G(z) \cdot w_{l,i}^t \,,
\]
where $w_{l,i}^t \in \R^M$ is the vectorized version of
parameters $w_{l,i,j}^t$, matrix $G(z) \in \R^{O \times M}$ ($O$ is the number of pixels
in $z$) is given as
\begin{equation}\label{gmtx}
\hspace*{-0.25cm}
\small
\underbrace{
\begin{bmatrix}
\varphi(\frac {|z_1 - \mu_1|}{\gamma}) & \varphi(\frac {|z_1 - \mu_2|}{\gamma}) &\cdots
&\varphi(\frac {|z_1 - \mu_M|}{\gamma})\\
\varphi(\frac {|z_2 - \mu_1|}{\gamma}) & \varphi(\frac {|z_2 - \mu_2|}{\gamma}) &\cdots
&\varphi(\frac {|z_2 - \mu_M|}{\gamma})\\
\vdots & \vdots & \ddots &\vdots\\
\varphi(\frac {|z_O- \mu_1|}{\gamma}) & \varphi(\frac {|z_O - \mu_2|}{\gamma}) &\cdots
&\varphi(\frac {|z_O  - \mu_M|}{\gamma})
\end{bmatrix}
}_\text{$G(z)$}
\underbrace
{\left[ \begin{array}{c} w_{l,i,1} \\ w_{l,i,2} \\ \vdots \\ w_{l,i,M} \end{array} \right]
}_\text{$w_{l,i}$} =
\underbrace
{\left[ \begin{array}{c} \phi_{l,i}\left(z_1\right) \\ \phi_{l,i}\left(z_2\right) \\ \vdots \\ \phi_{l,i}\left(z_O\right) \end{array} \right]
}_\text{$\phi_{l,i}(z)$} \,.
\end{equation}
Then, we can obtain the desired gradients of $u_t$ with respect to
the influence function $\phi_{l,i}^t$, given as
\begin{equation}\label{grad_uw}
\frac {\partial u_t}{\partial {\phi_{l,i}^t}} \triangleq
\frac {\partial u_t}{\partial {w_{l,i}^t}} = -G^\top {\bar{K_{l,i}^t}}^\top A_l\,.
\end{equation}

\subsubsection{Computation of the derivatives of $\frac {\partial u_{t}}
{\partial \lambda^t}$}
From the diffusion model \eqref{Mg-diffusion}, it is easy to check that
\begin{equation}
\frac {\partial u_{t}}{\partial \lambda^t} =
(u_{t-1} -f )^\top\,.
\end{equation}
\subsubsection{Computation of the derivatives of $\frac {\partial u_{t+1}}{\partial u_{t}}$}
As pointed out in \eqref{iterstep}, in the joint training phase,
we also need to compute $\frac {\partial u_{t+1}}{\partial u_{t}}$.
This gradient can be obtained from the diffusion equation \eqref{Mg-diffusion}.
The final formulation of $\frac {\partial u_{t+1}}{\partial u_{t}}$ is given as follows
\begin{equation}
\label{gradientfinal}
\frac {\partial u_{t+1}}{\partial u_{t}} = (1-\lambda^{t+1})\ident -
\suml{l = 0}{L}\sum\limits_{i = 1}^{N_k}
A_l^\top \left({K_{l,i}^{t+1}}\right)^\top \cdot \Lambda_{l,i} \cdot
\left({\bar K_{l,i}^{t+1}}\right)^\top A_l\,,
\end{equation}
where $\Lambda_{l,i}$ is a diagonal matrix
$\Lambda_{l,i} = \text{diag}({\phi_{l,i}^{t+1}}'(z_1), \cdots,
{\phi_{l,i}^{t+1}}'(z_p))$ with $z = K_{l,i}^{t+1} A_l u_{t}$.
Remember that the matrices $K_{l,i}^{t+1}$ and $\bar K_{l,i}^{t+1}$ are related to
th linear kernels $k_{l,i}^{t+1}$ and $\bar k_{l,i}^{t+1}$, respectively.

\subsection{Gradients for the case of Poisson denoising}\label{sec:poisson}
\subsubsection{Deriving the diffusion process for Poisson denoising}
For the case of Poisson denoising, the data term should be chosen as
\begin{equation}\label{datatermpoisson}
\lambda \langle u-f \mathrm{log}u,1 \rangle\,,
\end{equation}
with $u > 0$. Casting it into the framework \eqref{multi-diffusion}, we have two choices:
(1) setting $\cD(u,f) = \lambda \langle u-f \mathrm{log}u,1 \rangle$ and
$\cG(u,f) = 0$; (2) setting
$\cD(u,f) = 0$ and
$\cG(u,f) = \lambda \langle u-f \mathrm{log}u,1 \rangle$.

Considering the first setup, we have $\phi(u,f) = \nabla_u \cD(u,f) =
\lambda (1-\frac{f}{u})$.
One can see that, this strategy is not applicable in practice,
because (a) it has an evident problem of numerical instability at the points
where $u$ is very close to zero; (b) this update rule can not guarantee that
the output image after one diffusion step is positive. Negative values of
$u$ will violate the constraint of the data term \eqref{datatermpoisson}.

Then we resort to the second strategy by setting
$\cD(u,f) = 0$ and
$\cG(u,f) = \lambda \langle u-f \mathrm{log}u,1 \rangle$.
The proximal mapping with respect to $\cG$ is given as the following minimization problem
\begin{equation}\label{subproblemGpoisson}
\left( \mat{I} + \partial \cG \right)^{-1}(\tilde{u}) = \arg\min\limits_{u} \frac{\|u - \tilde
{u}\|^2_2}{2} + \lambda \langle u-f \mathrm{log}u,1 \rangle \,.
\end{equation}
The solution of \eqref{subproblemGpoisson} is given by the following point-wise operation
\begin{equation}\label{subproblemGIdiv}
\hat u = \left( \mat{I} +  \partial \cG \right)^{-1}(\tilde{u})  = \frac{ \tilde{u} - \lambda
+\sqrt{\left( \tilde{u} - \lambda \right)^2+4 \lambda f}}{2} \,.
\end{equation}
Note that $\hat u$ is always positive if $f > 0$, i.e, this update rule can
guarantee $\hat u > 0$ in diffusion steps.

As a consequence, the diffusion process for Poisson denoising using the proximal gradient method is formulated as
\begin{equation}\label{diffusionprocessfinal}
u_{t+1}  = \frac{ \tilde{u}_{t+1} - \lambda^{t+1}
+\sqrt{\left( \tilde{u}_{t+1} - \lambda^{t+1} \right)^2+4 \lambda^{t+1} f}}{2} \,,
\end{equation}
where $\tilde{u}_{t+1} = u_{t} - \sum\limits_{l = 0}^{L} \sum\limits_{i = 1}^{N_k}
A_l^\top \left(\bar k_{l,i}^{t+1} * \phi_{l,i}^{t+1}(k_{l,i}^{t+1}
* A_l u_{t})\right)$.

\subsection{Computing the gradients for training}
Following the results presented in the previous subsection, for the training phase, we
mainly need to compute
$\frac {\partial u_{t+1}}{\partial u_{t}}$ and
$\frac {\partial u_t}{\partial \Theta_t}$.

Based on the diffusion procedure \eqref{diffusionprocessfinal},
$\frac{\partial u_{t+1}}{\partial u_t}$ is computed via the chain rule, given as
\begin{equation}\label{derv1}
\frac{\partial u_{t+1}}{\partial u_t} =\frac{\partial \tilde{u}_{t+1}}{\partial u_t} \cdot \frac{\partial u_{t+1}}{\partial \tilde{u}_{t+1}}.
\end{equation}

$\frac{\partial \tilde{u}_{t+1}}{\partial u_t}$ can be easily obtained by
following the result \eqref{gradientfinal}.
Concerning the part $\frac{\partial u_{t+1}}{\partial \tilde{u}_{t+1}}$, it
can be computed according to \eqref{diffusionprocessfinal} and is formulated as
\begin{equation}\label{derv2}
\frac{\partial u_{t+1}}{\partial \tilde{u}_{t+1}} =
\text{diag}(y_1, \cdots, y_N)\,,
\end{equation}
where $\{ y_i \}_{i=1}^{i=N}$ denote the elements of
\[
y = \frac{1}{2} \left[ 1+\frac{\tilde{u}_{t+1}-\lambda^{t+1}}
{\sqrt{\left( \tilde{u}_{t+1}-\lambda^{t+1} \right)^2+4\lambda^{t+1} f}} \right]\,.
\]

Concerning the gradients $\frac{\partial u_t}{\partial \Theta_t}$
($\Theta_t$ involves $\left\{ \lambda^t,\phi_{l,i}^t, k_{l,i}^t \right\}$),
it is worthy noting that the gradients of $u_t$ with respect to
$\left\{\phi_{l,i}^t, k_{l,i}^t \right\}$ are only associated with
\[
\tilde{u}_{t} = u_{t-1} - \sum\limits_{l = 0}^{L} \sum\limits_{i = 1}^{N_k}
A_l^\top \left(\bar k_{l,i}^t * \phi_{l,i}^t(k_{l,i}^t * A_l u_{t-1})\right)\,.
\]
Therefore, the gradient of $u_t$ with respect to $\phi_{l,i}^t$ and $k_{l,i}^t$
is computed via
\[
\frac{\partial u_t}{\partial \phi_{l,i}^t} = \frac{\partial \tilde{u}_t}{\partial \phi_{l,i}^t} \cdot  \frac{\partial u_t}{\partial \tilde{u}_t},
\,\mathrm{and}\,\,
\frac{\partial u_t}{\partial k_{l,i}^t} =  \frac{\partial \tilde{u}_t}{\partial k_{l,i}^t} \cdot \frac{\partial u_t}{\partial \tilde{u}_t},
\]
where the derivations of $\frac{\partial \tilde{u}_t}{\partial \phi_{l,i}^t}$ and $\frac{\partial \tilde{u}_t}{\partial k_{l,i}^t}$ are given as \eqref{grad_uw} and
\eqref{grad_uk}, respectively.
The gradients of $\frac{\partial u_t}{\partial \tilde{u}_t}$ are calculated
similar to \eqref{derv2}.

The gradient of $u_t$ with respect to $\lambda^t$ is computed as
\begin{equation}\label{derv4}
\frac{\partial u_t}{\partial \lambda^t} =
(z_1, \cdots, z_N)\,,
\end{equation}
where $\{ z_i \}_{i=1}^{i=N}$ denote the elements of
\[
z = \frac{1}{2} \left[ -1+\frac{\left(\lambda^t-\tilde{u}_t\right)+2f}{\sqrt{\left( \tilde{u}_t-\lambda^t \right)^2+4\lambda^t f}} \right]\,.
\]
Note that $\frac{\partial u_t}{\partial \lambda^t}$ is written as
a row vector.

\section{Experiments}
In this section, we employed the fully Trained Multi-scale Nonlinear Diffusion Model
(MSND) for Gaussian denoising and Poisson Denoising. The corresponding nonlinear diffusion process of stage $T$ with filters of size $m\times m$ is expressed as $\mathrm{MSND}_{m \times m}^T$. Note that there are different options for the number of scales. It is not a stretch to infer that more scales leads to better denoising performance, but increases the cost of computing. Without loss of generality, in this study we employed three scales with the scale factors 1.5, 2 and 3 respectively. The number of filters for each scale is $\left( m^2-1 \right)$ in each stage, if not specified.

To generate the training data for our denoising experiments, we cropped a 180$\times$180 pixel region
from each image of the Berkeley segmentation
dataset \cite{MartinFTM01}, resulting in a total of 400 training samples of size 180 $\times$ 180. This setting is the same as TNRD \cite{chenCVPR15} for fair evaluation. After training the models, we evaluated them on 68 test images originally introduced by
\cite{roth2009fields}, which have since become a reference set for image
denoising. To provide a comprehensive comparison, the standard deviations $\sigma$ for the Gaussian noise are set as 25, 50, 75 and 100, respectively. Meanwhile, the peak values of Poisson noise are distributed between 1 to 4. For saving the training time, the $\mathrm{MSND}_{7 \times 7}^5$ model is employed. Meanwhile, we also tested the $\mathrm{MSND}_{5 \times 5}^5$ model for comparison. Note that, the diffusion model needs to be trained respectively for different noise levels and different noise types.

\begin{figure}[t!]
\centering
\subfigure[\scriptsize Clean Image $I_0$]{
\includegraphics[width=0.17\textwidth]{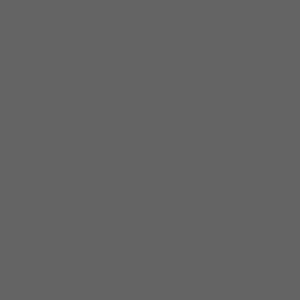}
}
\subfigure[\scriptsize Noisy image $I_m$. $\sigma=50$]{
\includegraphics[width=0.17\textwidth]{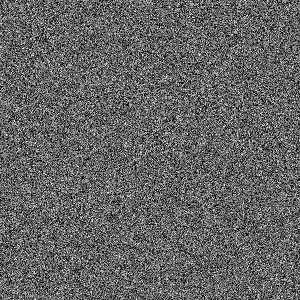}
}
\subfigure[\scriptsize The under-sampled noisy image $I_m^{1.5}$.]{
\includegraphics[width=0.17\textwidth]{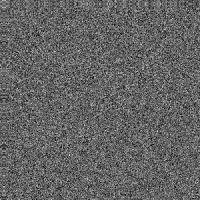}
}
\subfigure[\scriptsize The under-sampled noisy image $I_m^{2}$.]{
\includegraphics[width=0.17\textwidth]{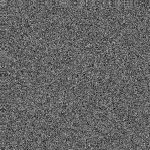}
}
\subfigure[\scriptsize The under-sampled noisy image $I_m^{3}$.]{
\includegraphics[width=0.17\textwidth]{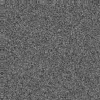}
}
\\
\subfigure[\scriptsize $\mathrm{TNRD}_{5 \times 5}^5$ (37.46/0.952)]{
\includegraphics[width=0.17\textwidth]{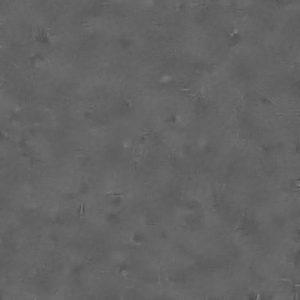}
}
\subfigure[\scriptsize $\mathrm{TNRD}_{7 \times 7}^5$ (39.22/0.970)]{
\includegraphics[width=0.17\textwidth]{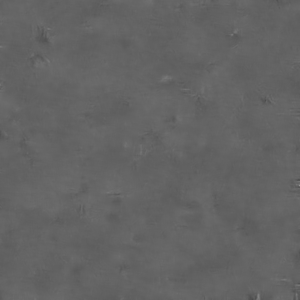}
}
\subfigure[\scriptsize $\mathrm{MSND}_{5 \times 5}^5$ (41.34/0.976)]{
\includegraphics[width=0.17\textwidth]{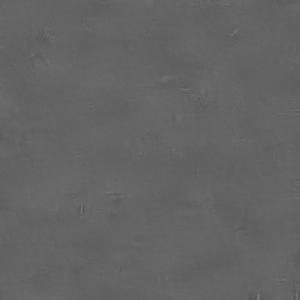}
}
\subfigure[\scriptsize $\mathrm{MSND}_{7 \times 7}^5$ (\textbf{43.64}/\textbf{0.985})]{
\includegraphics[width=0.17\textwidth]{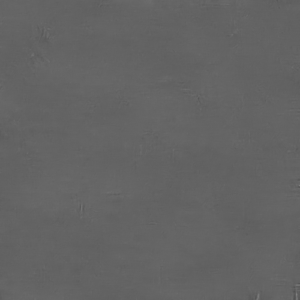}
}
\caption{Denoising results comparison. The results are reported by PSNR/SSIM index. Best results are marked. Note that the under-sampled noisy image $I_m^{n}$ is obtained by sub-sampling $I_m$ with scale factor $n$. The resulting standard deviation is $\sigma=\frac{50}{n}$.}
\label{veryfy}
\end{figure}

For Gaussian denoising task, the proposed algorithm is compared with five representative state-of-the-art methods: BM3D \cite{dabov2007image}, WNNM 
\cite{gu2014weighted}, EPLL-GMM \cite{zoran2011learning}, multi-scale EPLL-GMM(MSEPLL) \cite{papyan2016multi} and TNRD \cite{chenCVPR15}. For Poisson noise denoising, we compared with the NLSPCA \cite{salmon2014poisson} and BM3D-based methods with the exact unbiased inverse Anscombe
\cite{INAnscombe5}, both with and without binning technique.
The corresponding codes are downloaded from the authors' homepage, and we used them as is.
Especially, for the binning technique in the Poisson denoising task, we closely followed \cite{salmon2014poisson} and
use a $3 \times 3$ ones kernel to increase the peak value to be 9 times higher, and
a bilinear interpolation for the upscaling of the low-resolution
recovered image.
Two commonly used quality measures are taken to evaluate the denoising performance, \ie, PSNR and
the structural similarity index (SSIM) \cite{ssim}. The range of SSIM
value lies in [0, 1] with 1 standing for perfect quality. The PSNR and SSIM values in the following subsections are evaluated by averaging denoised results of 68 test images. Note that the PSNR value for Poisson noise denoising is computed based on the peak value, instead of 255 for Gaussian noise denosing task on 8-bit images.

\subsection{Gaussian Denoising}
The main motivation for constructing
a multi-scale model was to remedy the artifacts obtained from
local processing of the image. Therefore, we begin with a preliminary test in Fig.~\ref{veryfy} which presents the denoised images
obtained by the local model TNRD \cite{chenCVPR15} and the proposed model, showing that indeed artifacts are better treated. In detail, by observing on the recovered images, we can see that the single-scale local model TNRD introduces some incorrect image features. This is because that local models infer the underlying structure solely from the local neighborhoods, which is heavily distorted by noise. In this case,
the advantage of multi-scale structure comes through.

The downsampling operation in the proposed multi-scale model MSND can cause the uncorrelated values of the noise to
become smaller and consequently make image structures
more visible, as shown in Fig.~\ref{veryfy}(c), (d) and (e). Moreover, down-sampling the image before denoising amounts to
enlarging the size of the neighborhood on which the denoising is performed, thus permitting to exploit larger-scale
information. As a consequence, the proposed multi-scale model MSND leads to great performance improvements by comparing Fig.~\ref{veryfy}(f,g) with (h,i).

\begin{figure*}[htbp]
\centering
\subfigure[\scriptsize Clean Image]{
\centering
\includegraphics[width=0.17\textwidth]{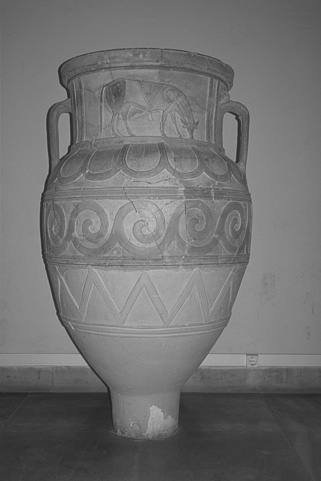}
}
\subfigure[\scriptsize Noisy image. $\sigma=50$]{
\centering
\includegraphics[width=0.17\textwidth]{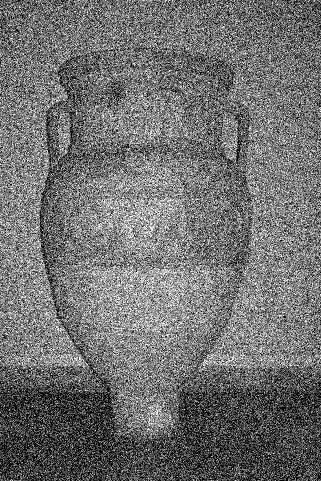}
}
\subfigure[\scriptsize BM3D(30.71/0.790)]{
\centering
\includegraphics[width=0.17\textwidth]{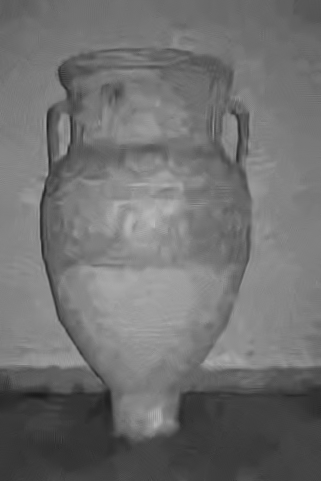}
}
\subfigure[\scriptsize EPLL(30.12/0.763)]{
\centering
\includegraphics[width=0.17\textwidth]{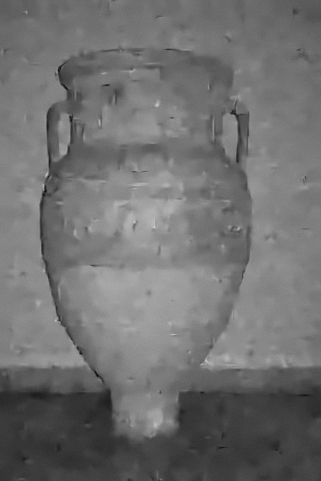}
}
\subfigure[\scriptsize WSEPLL(30.87/0.794)]{
\centering
\includegraphics[width=0.17\textwidth]{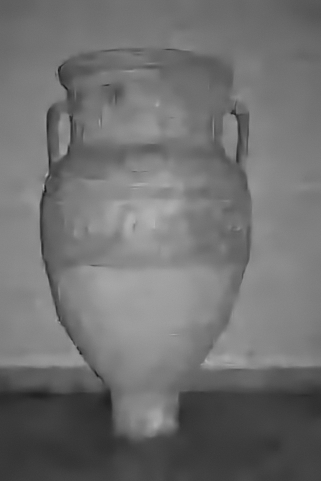}
}
\subfigure[\scriptsize WNNM(30.68/0.790)]{
\centering
\includegraphics[width=0.17\textwidth]{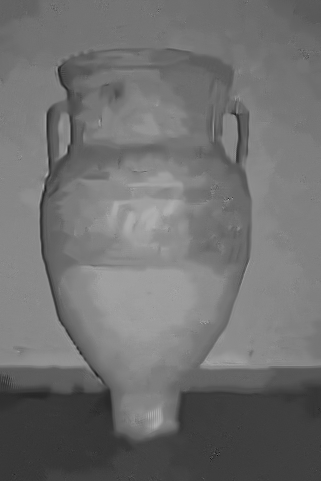}
}
\subfigure[\scriptsize $\mathrm{TNRD}_{5 \times 5}^5$ (30.38/0.775)]{
\centering
\includegraphics[width=0.17\textwidth]{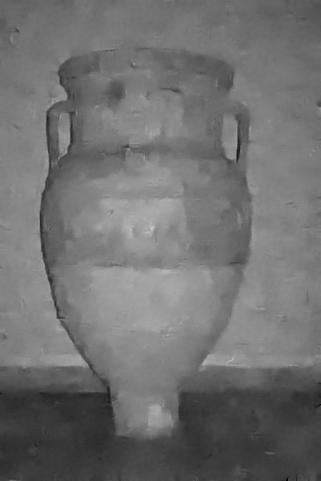}
}
\subfigure[\scriptsize $\mathrm{TNRD}_{7 \times 7}^5$ (30.63/0.785)]{
\centering
\includegraphics[width=0.17\textwidth]{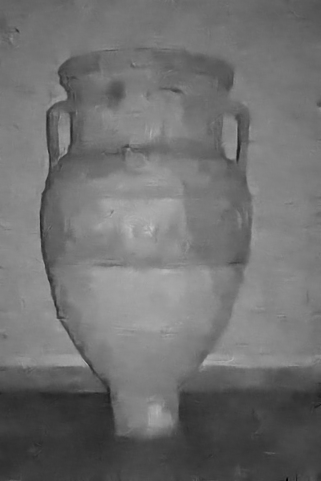}
}
\subfigure[\scriptsize $\mathrm{MSND}_{5 \times 5}^5$ (30.79/0.789)]{
\centering
\includegraphics[width=0.17\textwidth]{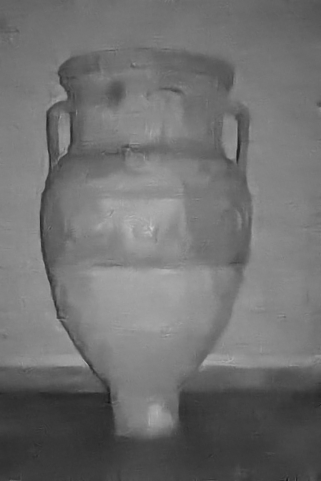}
}
\subfigure[\scriptsize $\mathrm{MSND}_{7 \times 7}^5$ (\textbf{30.95}/\textbf{0.794})]{
\centering
\includegraphics[width=0.17\textwidth]{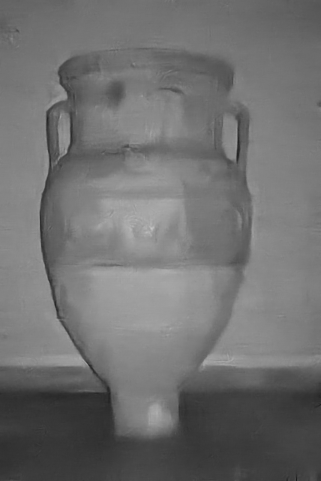}
}
\caption{Gaussian noise denoising results comparison. The results are reported by PSNR/SSIM index. Best results are marked.}
\label{sigma50}
\end{figure*}

\begin{figure*}[htbp]
\centering
\subfigure[\scriptsize Clean Image]{
\centering
\includegraphics[width=0.17\textwidth]{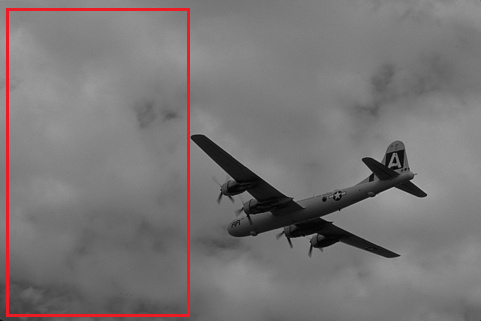}
}
\subfigure[\scriptsize Noisy image. $\sigma=75$]{
\centering
\includegraphics[width=0.17\textwidth]{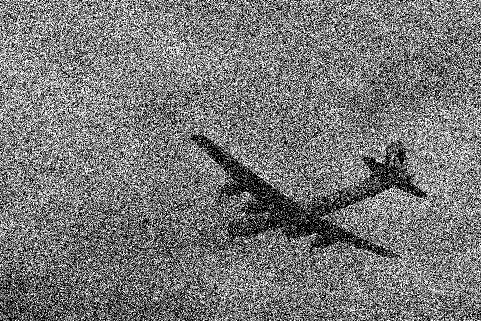}
}
\subfigure[\scriptsize BM3D(31.32/0.899)]{
\centering
\includegraphics[width=0.17\textwidth]{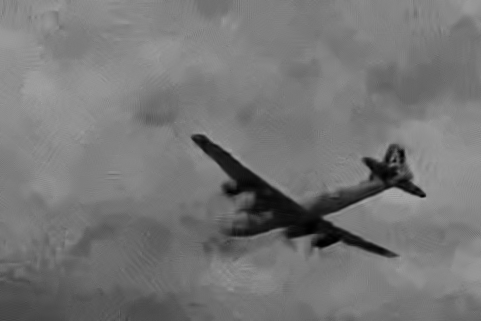}
}
\subfigure[\scriptsize EPLL(30.14/0.831)]{
\centering
\includegraphics[width=0.17\textwidth]{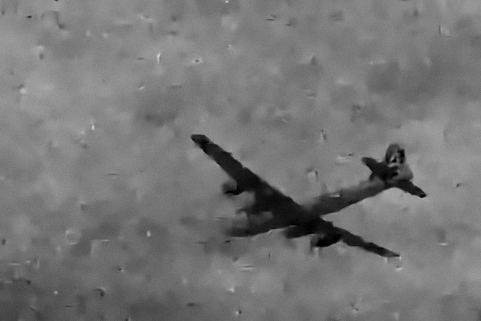}
}
\subfigure[\scriptsize WSEPLL(32.05/0.927)]{
\centering
\includegraphics[width=0.17\textwidth]{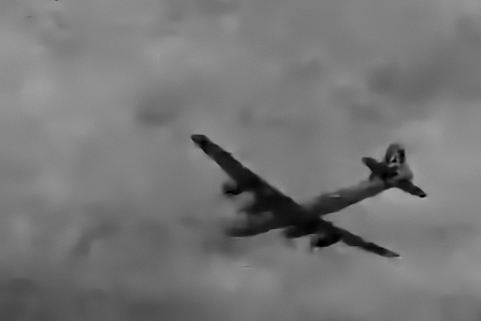}
}
\subfigure[\scriptsize WNNM(31.53/0.913)]{
\centering
\includegraphics[width=0.17\textwidth]{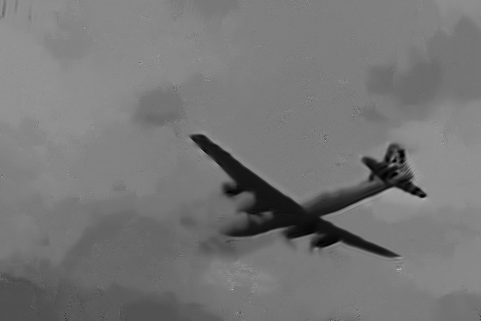}
}
\subfigure[\scriptsize $\mathrm{TNRD}_{5 \times 5}^5$ (31.28/0.894)]{
\centering
\includegraphics[width=0.17\textwidth]{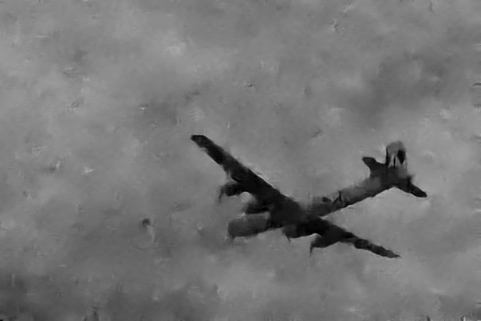}
}
\subfigure[\scriptsize $\mathrm{TNRD}_{7 \times 7}^5$ (31.85/0.911)]{
\centering
\includegraphics[width=0.17\textwidth]{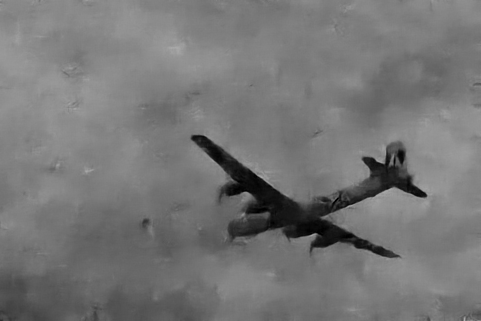}
}
\subfigure[\scriptsize $\mathrm{MSND}_{5 \times 5}^5$ (32.31/0.919)]{
\centering
\includegraphics[width=0.17\textwidth]{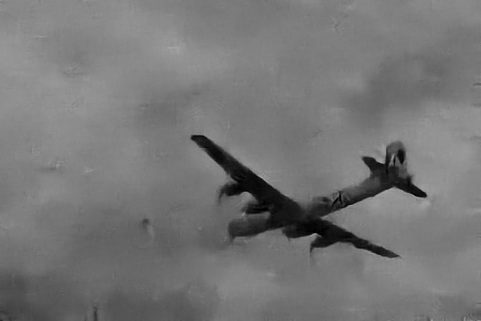}
}
\subfigure[\scriptsize $\mathrm{MSND}_{7 \times 7}^5$ (\textbf{32.61}/\textbf{0.928})]{
\centering
\includegraphics[width=0.17\textwidth]{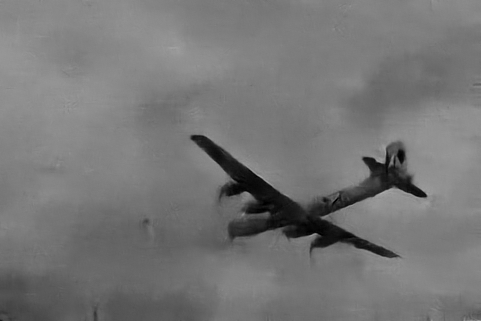}
}
\caption{Gaussian noise denoising results comparison. The results are reported by PSNR/SSIM index. Best results are marked.}
\label{sigma75}
\end{figure*}

\begin{figure*}[htbp]
\centering
\subfigure[\scriptsize Clean Image]{
\centering
\includegraphics[width=0.17\textwidth]{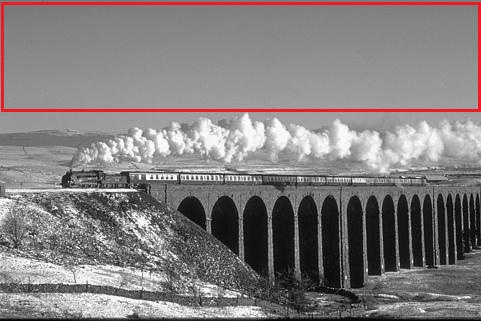}
}
\subfigure[\scriptsize Noisy image. $\sigma=100$]{
\centering
\includegraphics[width=0.17\textwidth]{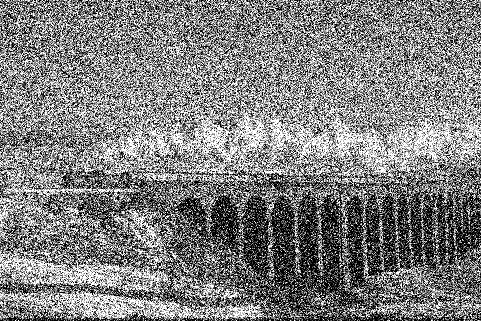}
}
\subfigure[\scriptsize BM3D(21.85/0.650)]{
\centering
\includegraphics[width=0.17\textwidth]{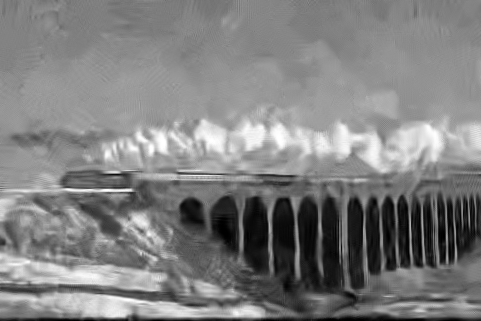}
}
\subfigure[\scriptsize EPLL(21.54/0.594)]{
\centering
\includegraphics[width=0.17\textwidth]{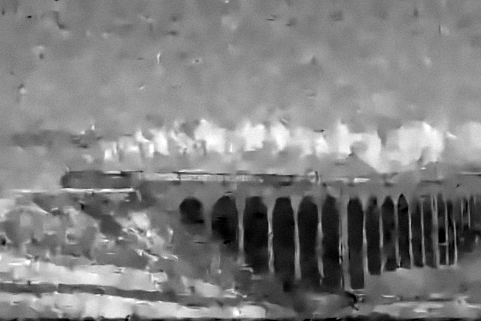}
}
\subfigure[\scriptsize WSEPLL(21.70/0.641)]{
\centering
\includegraphics[width=0.17\textwidth]{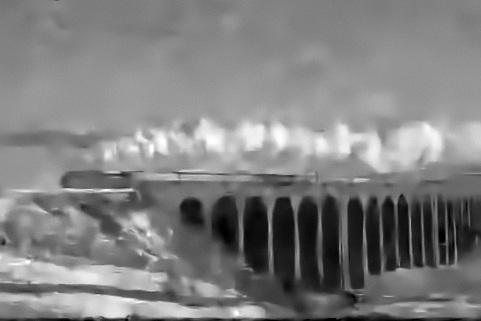}
}
\subfigure[\scriptsize WNNM(\textbf{22.08}/0.671)]{
\centering
\includegraphics[width=0.17\textwidth]{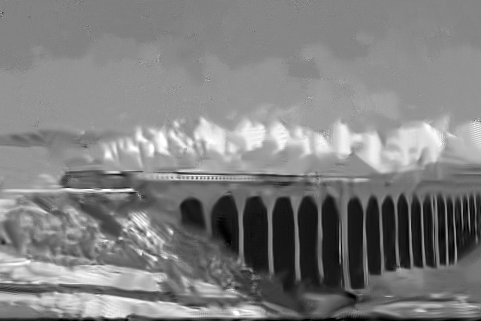}
}
\subfigure[\scriptsize $\mathrm{TNRD}_{5 \times 5}^5$ (21.67/0.636)]{
\centering
\includegraphics[width=0.17\textwidth]{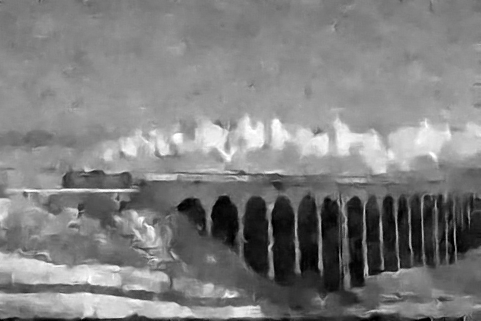}
}
\subfigure[\scriptsize $\mathrm{TNRD}_{7 \times 7}^5$ (21.87/0.659)]{
\centering
\includegraphics[width=0.17\textwidth]{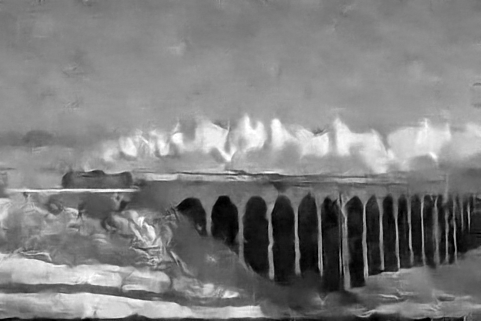}
}
\subfigure[\scriptsize $\mathrm{MSND}_{5 \times 5}^5$ (22.01/0.671)]{
\centering
\includegraphics[width=0.17\textwidth]{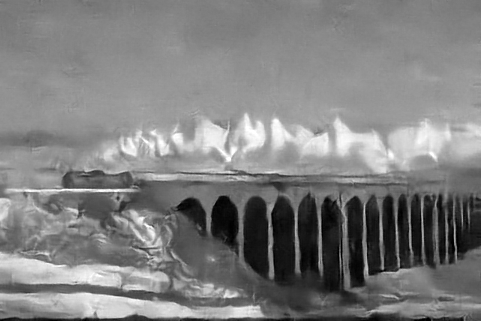}
}
\subfigure[\scriptsize $\mathrm{MSND}_{7 \times 7}^5$ (22.07/\textbf{0.683})]{
\centering
\includegraphics[width=0.17\textwidth]{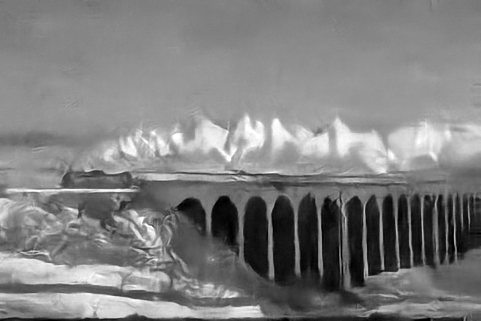}
}
\caption{Gaussian noise denoising results comparison. The results are reported by PSNR/SSIM index. Best results are marked.}
\label{sigma100}
\end{figure*}

\begin{table}[t!]
\footnotesize
\centering
\begin{tabular}{|c |c |c |c |c|}
\hline
Method & $\sigma=25$ &$\sigma=50$ &$\sigma=75$ &$\sigma=100$\\
\hline
\hline
BM3D & 28.57/0.802 &25.62/0.687&24.21/0.622&23.24/0.576\\
\hline
EPLL & 28.68/0.812 &25.68/0.688&24.10/0.606&23.06/ 0.547\\
\hline
MSEPLL & 28.78/0.812 &25.83/0.692&24.30/0.620&23.32/0.573\\
\hline
WNNM & 28.80/0.808 &25.83/0.698&24.39/ 0.626&23.39/0.585\\
\hline
$\mathrm{TNRD}_{5 \times 5}^5$& 28.77/0.808 &25.83/0.692&24.33/0.624&23.30/0.574\\
\hline
$\mathrm{TNRD}_{7 \times 7}^5$& 28.93/0.816 &25.96/0.702&24.45/0.632&23.47/0.586\\
\hline
$\mathrm{MSND}_{5 \times 5}^5$& 28.92/0.815 &26.01/0.704&24.54/0.638&23.58/0.593\\
\hline
$\mathrm{MSND}_{7 \times 7}^5$& \textbf{28.98}/\textbf{0.820} &\textbf{26.06}/\textbf{0.709}&\textbf{24.58}/\textbf{0.642}&\textbf{23.62}/\textbf{0.598}\\
\hline
\end{tabular}
\caption{Comparison of the performance on the Gaussian noise denoising of the test algorithms in terms of PSNR and SSIM. Best results are marked.}
\label{resultshow1}
\end{table}

\begin{figure}[t!]
\centering
\subfigure[$\sigma=50$, the filter size is $5\times 5$]{
\includegraphics[width=0.4\textwidth]{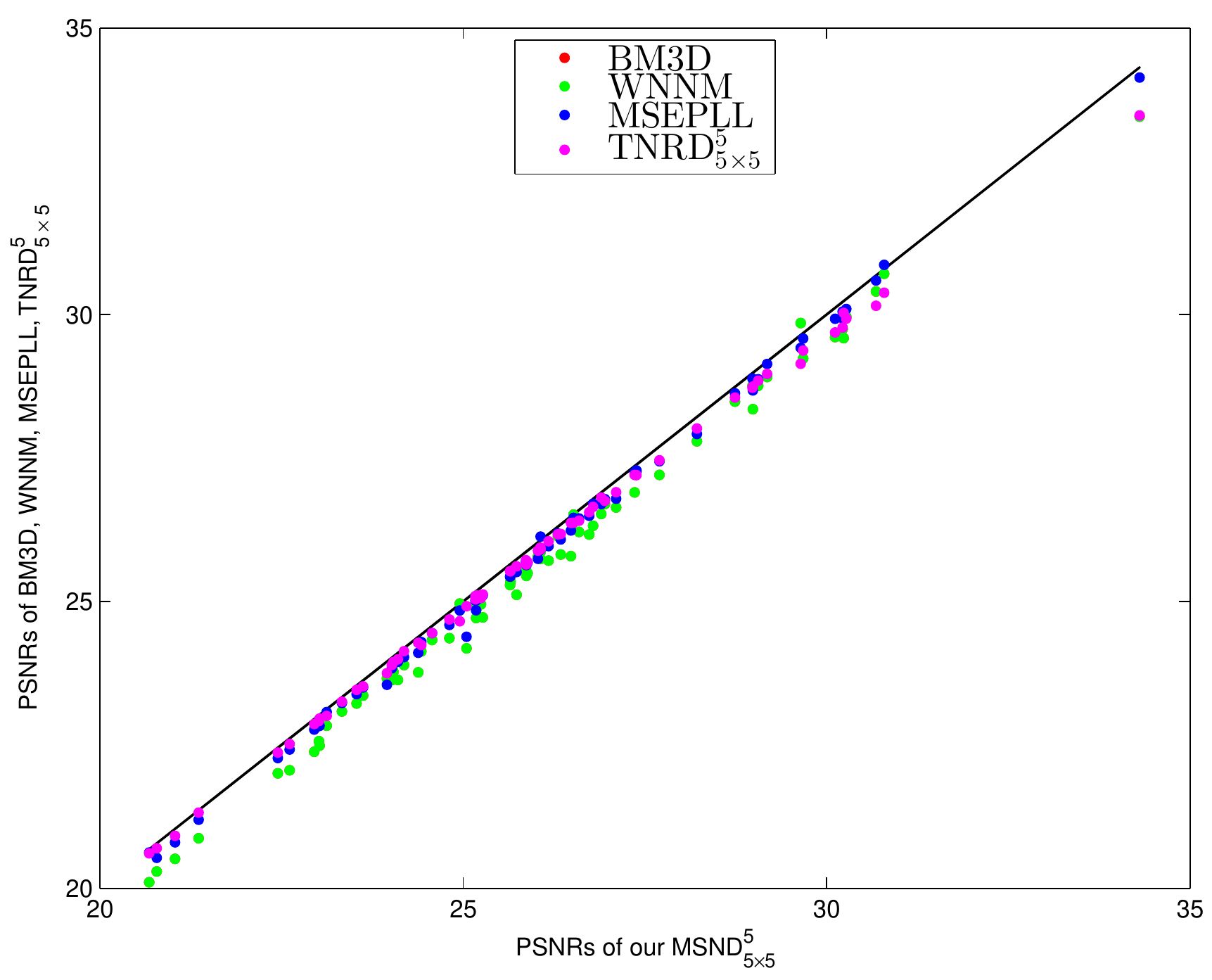}
}
\subfigure[$\sigma=75$, the filter size is $5\times 5$]{
\includegraphics[width=0.4\textwidth]{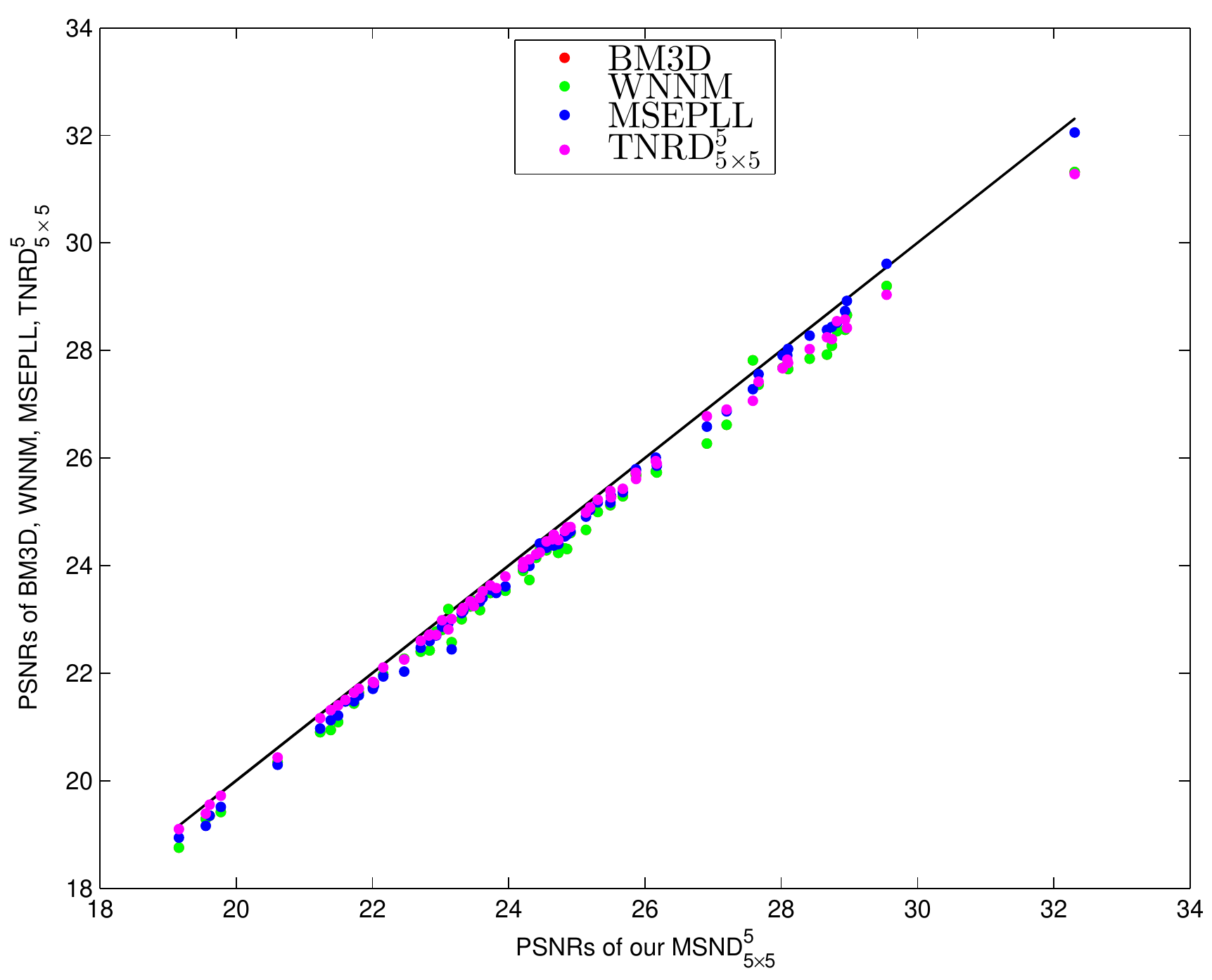}
}\\
\subfigure[$\sigma=50$, the filter size is $7\times 7$]{
\includegraphics[width=0.4\textwidth]{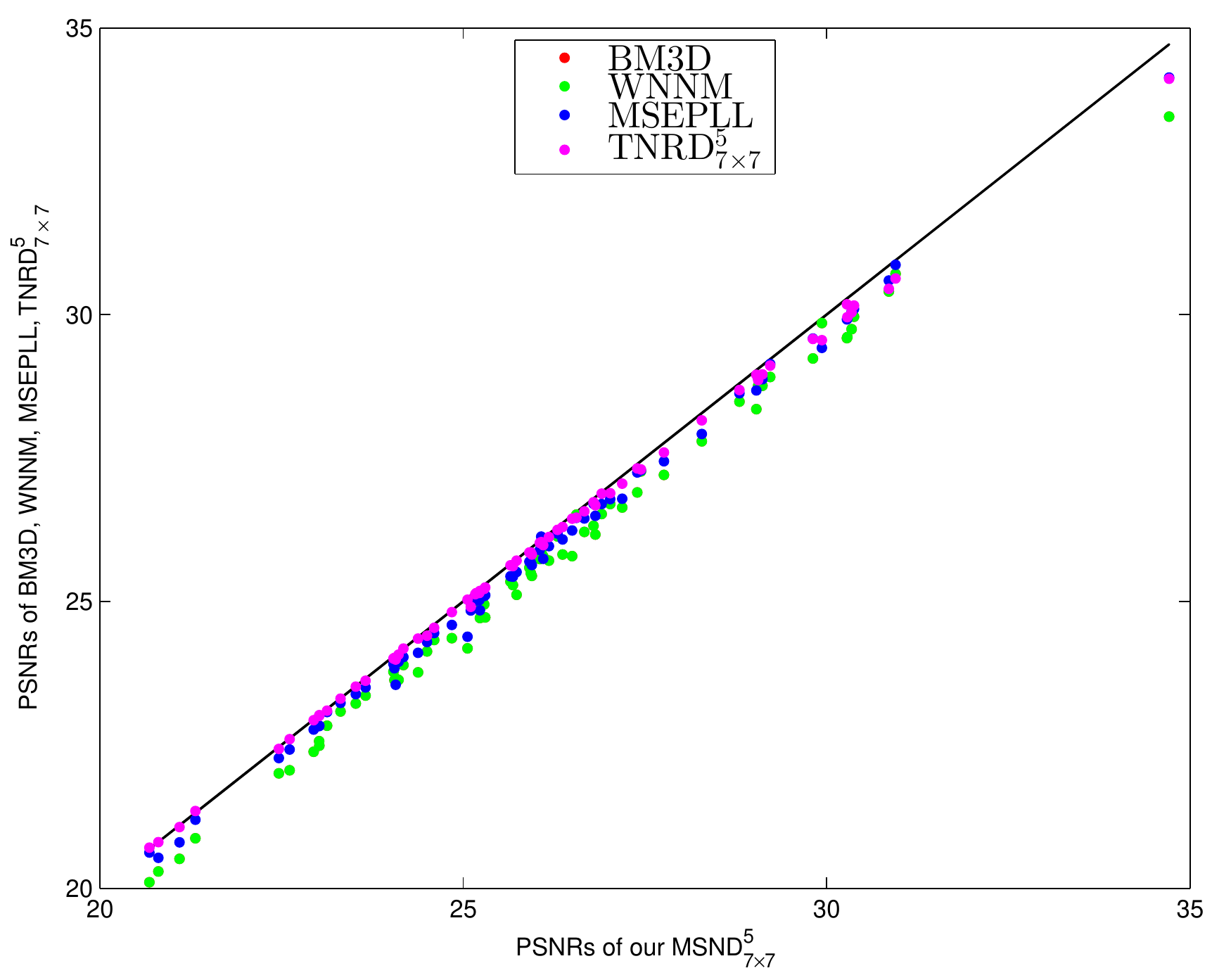}
}
\subfigure[$\sigma=75$, the filter size is $7\times 7$]{
\includegraphics[width=0.4\textwidth]{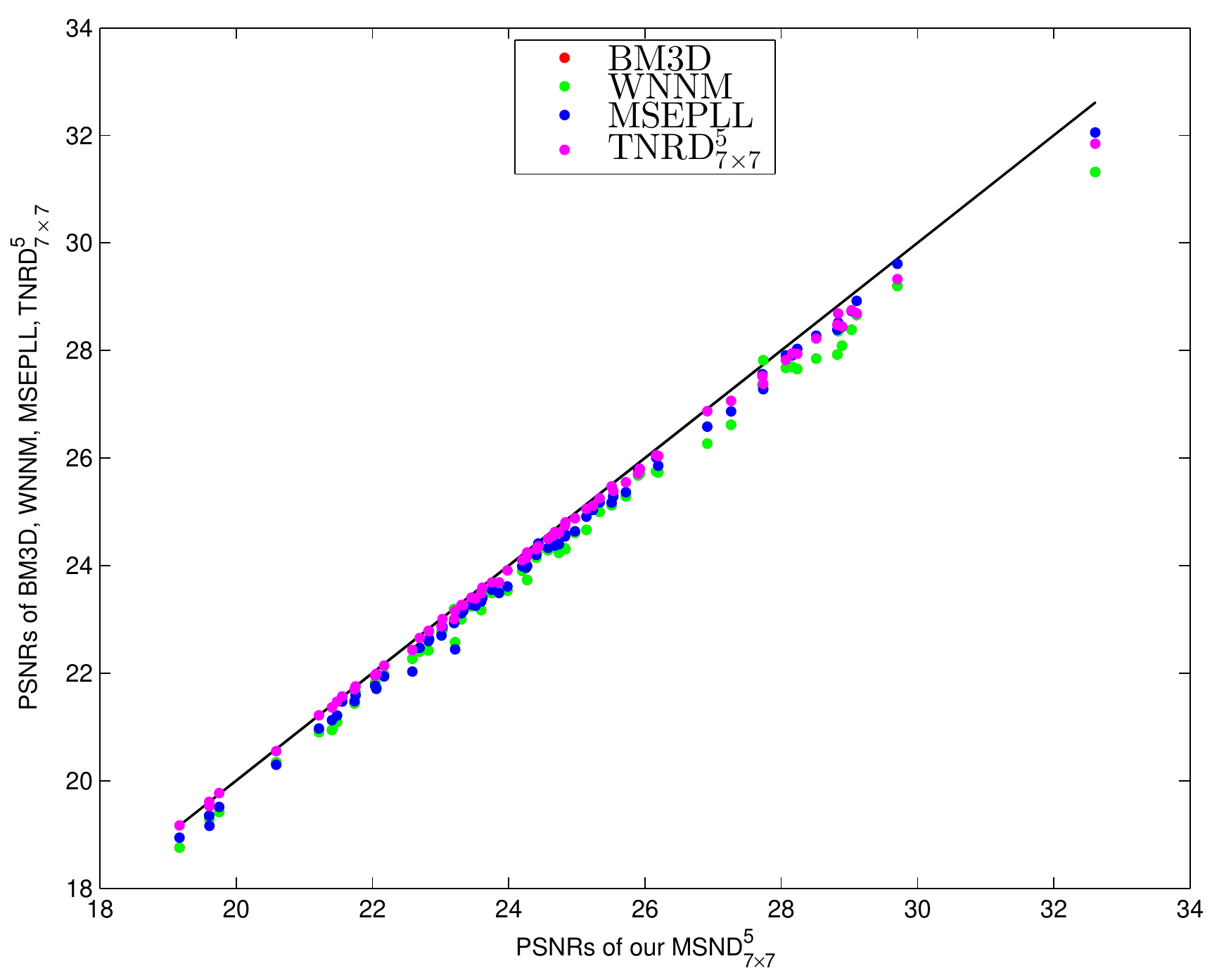}
}
\caption{Scatter plot of the PSNRs over 68 test images produced by the proposed MSND model, BM3D, WNNM, MSEPLL and TNRD. A point above the diagonal
line (i.e., $y = x$) means a better performance than the proposed model, and a
point below this line indicates a inferior result. One can see that the proposed MSND model outperforms the other test methods on most of the images.}
\label{plotcompare}
\end{figure}

Examining the recovery images in Fig.~\ref{sigma50}-Fig.~\ref{sigma100}, the nonlocal technique BM3D is affected by the structured signal-like artifacts that appear in homogeneous areas of the image. This phenomenon is originated from the selection process of similar image patches in the BM3D denoising scheme. The selection process is easily influenced by the noise itself, especially in flat areas of the image, which can be dangerously self-referential. Meanwhile, we find that the method TNRD introduces block-type artifacts, whose reason has been explained above.
In comparison with TNRD, the proposed algorithm MSND is more powerful on suppressing artifacts and recovering the homogeneous
regions of the image, especially within the red rectangles in the presented figures.

The recovery errors in terms of PSNR (in dB) and SSIM are
summarized in Table~\ref{resultshow1}. Comparing the indexes in Table~\ref{resultshow1}
and the denoising results in the present figures, the best
overall performance is provided by the proposed method MSND. Especially, with the noise level increasing, the superiority of MSND enhances as well. This is natural since the artifacts brought by the single-scale local model are more serious when the noise level is higher. Meanwhile, the proposed model not only leads to an improvement
in smooth regions, but also performs at least as well as the single-scale local model in highly textured areas.

Fig.~\ref{plotcompare} presents a detailed comparison between our learned MSND model and four state-of-the-art methods over 68 natural
images for noise level $\sigma$ = 50 and $\sigma$ = 75. One can see that our proposed MSND outperforms the compared denoising methods for most images.


\subsection{Poisson Denoising}
\begin{figure*}[t!]
\centering
\subfigure[\scriptsize Clean Image]{
\centering
\includegraphics[width=0.17\textwidth]{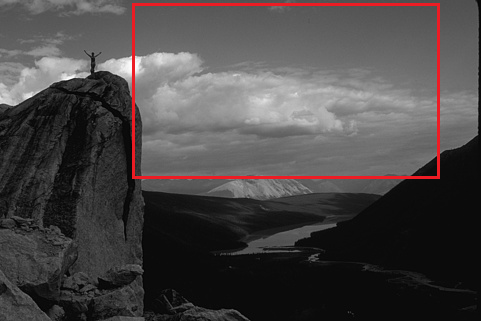}
}
\subfigure[\scriptsize Noisy image. Peak=1]{
\centering
\includegraphics[width=0.17\textwidth]{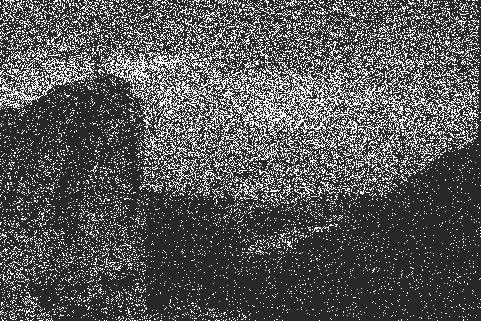}
}
\subfigure[\scriptsize NLSPCA (25.78/0.717)]{
\centering
\includegraphics[width=0.17\textwidth]{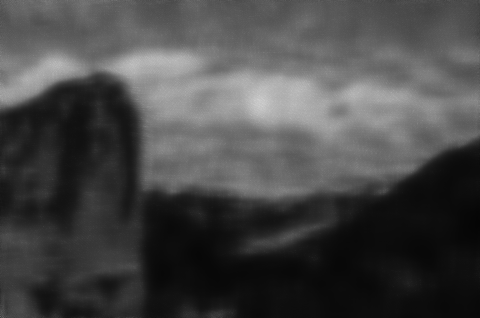}
}
\subfigure[\scriptsize NLSPCAbin (23.72/0.715)]{
\centering
\includegraphics[width=0.17\textwidth]{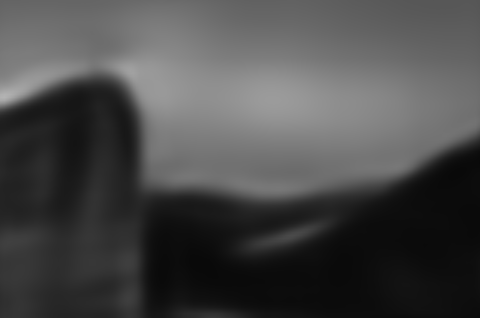}
}
\subfigure[\scriptsize BM3D (25.51/0.714)]{
\centering
\includegraphics[width=0.17\textwidth]{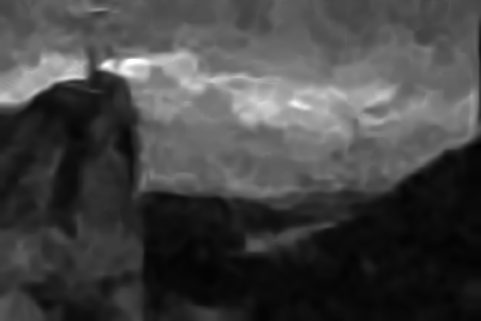}
}\\
\subfigure[\scriptsize BM3Dbin (26.22/0.738)]{
\centering
\includegraphics[width=0.17\textwidth]{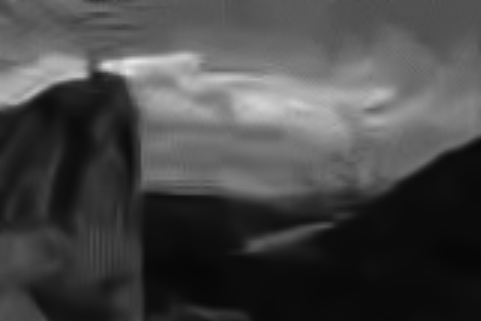}
}
\subfigure[\scriptsize $\mathrm{TNRD}_{5 \times 5}^5$ (26.02/0.675)]{
\centering
\includegraphics[width=0.17\textwidth]{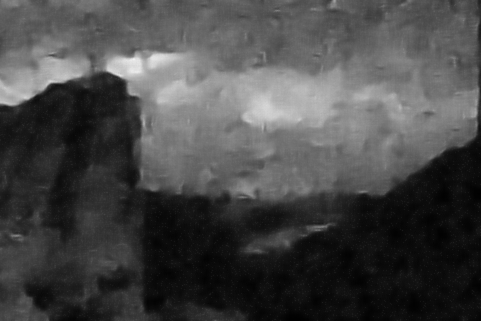}
}
\subfigure[\scriptsize $\mathrm{TNRD}_{7 \times 7}^5$ (26.29/0.690)]{
\centering
\includegraphics[width=0.17\textwidth]{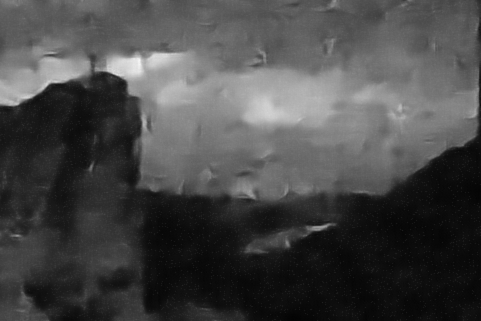}
}
\subfigure[\scriptsize $\mathrm{MSND}_{5 \times 5}^5$ (26.47/0.706)]{
\centering
\includegraphics[width=0.17\textwidth]{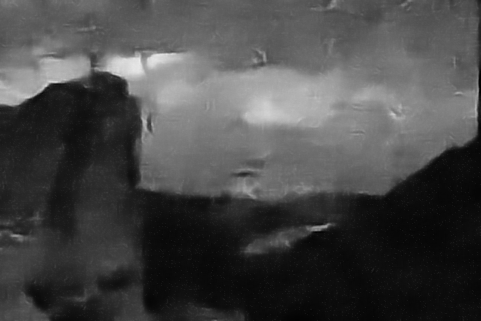}
}
\subfigure[\scriptsize $\mathrm{MSND}_{7 \times 7}^5$ (\textbf{26.65}/\textbf{0.714})]{
\centering
\includegraphics[width=0.17\textwidth]{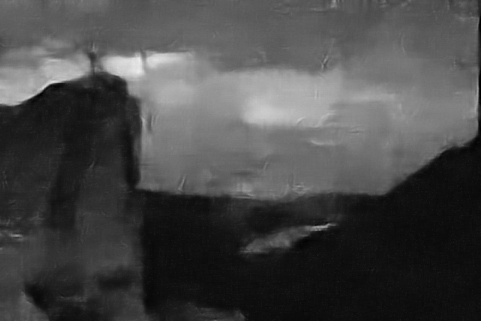}
}
\caption{Poisson noise denoising results comparison. The results are reported by PSNR/SSIM index. Best results are marked.}
\label{peak1}
\end{figure*}

\begin{figure*}[htbp]
\centering
\subfigure[\scriptsize Clean Image]{
\centering
\includegraphics[width=0.17\textwidth]{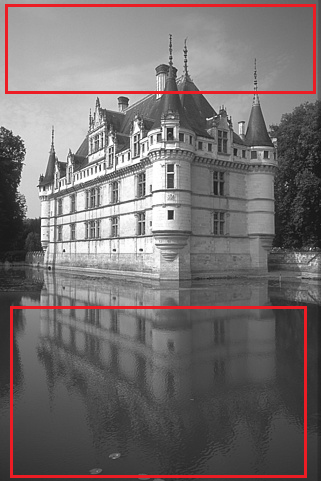}
}
\subfigure[\scriptsize Noisy image. Peak=2]{
\centering
\includegraphics[width=0.17\textwidth]{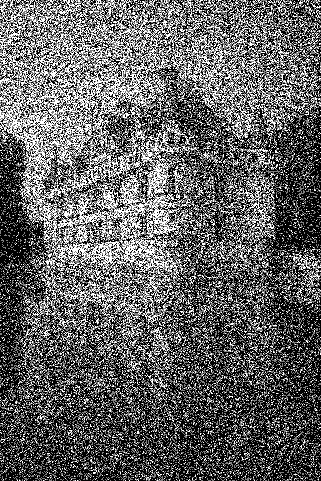}
}
\subfigure[\scriptsize NLSPCA (21.11/0.613)]{
\centering
\includegraphics[width=0.17\textwidth]{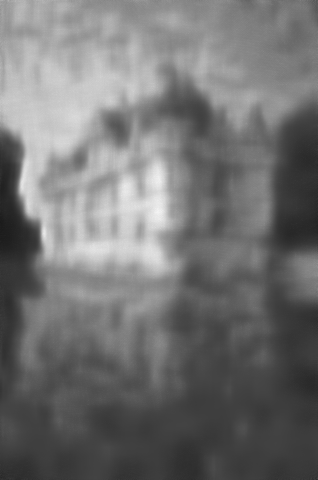}
}
\subfigure[\scriptsize NLSPCAbin (20.30/0.606)]{
\centering
\includegraphics[width=0.17\textwidth]{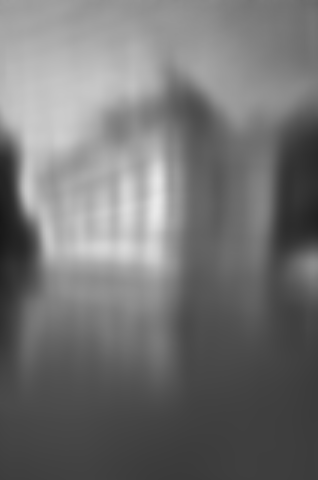}
}
\subfigure[\scriptsize BM3D (22.03/0.627)]{
\centering
\includegraphics[width=0.17\textwidth]{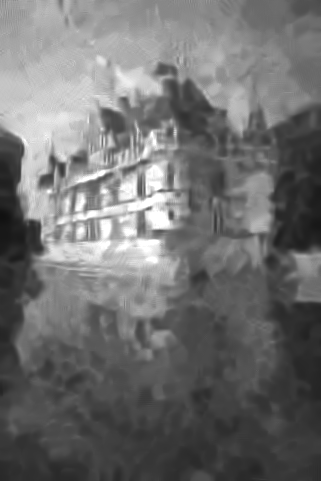}
}\\
\subfigure[\scriptsize BM3Dbin (21.99/0.643)]{
\centering
\includegraphics[width=0.17\textwidth]{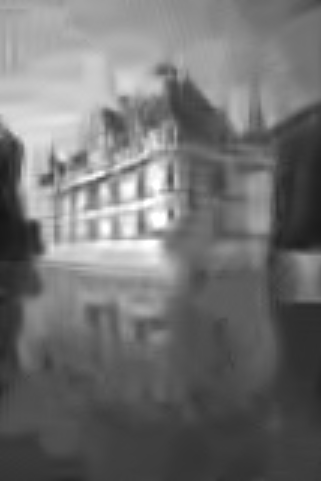}
}
\subfigure[\scriptsize $\mathrm{TNRD}_{5 \times 5}^5$ (22.40/0.623)]{
\centering
\includegraphics[width=0.17\textwidth]{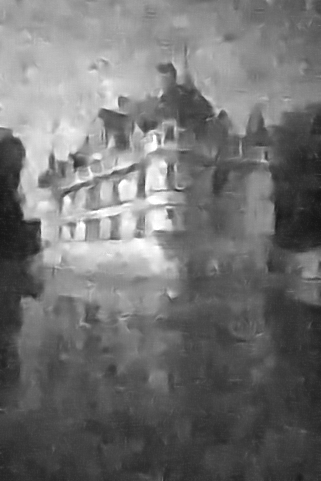}
}
\subfigure[\scriptsize $\mathrm{TNRD}_{7 \times 7}^5$ (22.51/0.640)]{
\centering
\includegraphics[width=0.17\textwidth]{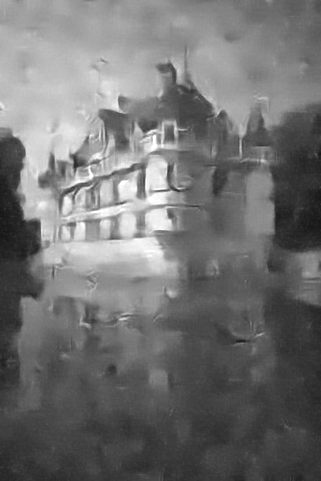}
}
\subfigure[\scriptsize $\mathrm{MSND}_{5 \times 5}^5$ (22.77/0.656)]{
\centering
\includegraphics[width=0.17\textwidth]{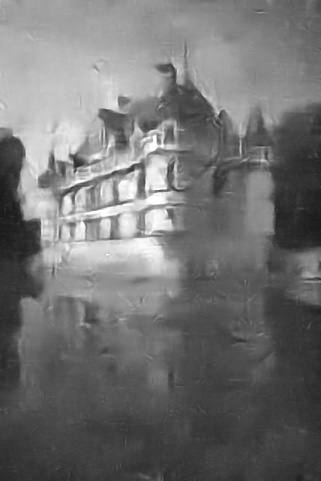}
}
\subfigure[\scriptsize $\mathrm{MSND}_{7 \times 7}^5$ (\textbf{22.84}/\textbf{0.662})]{
\centering
\includegraphics[width=0.17\textwidth]{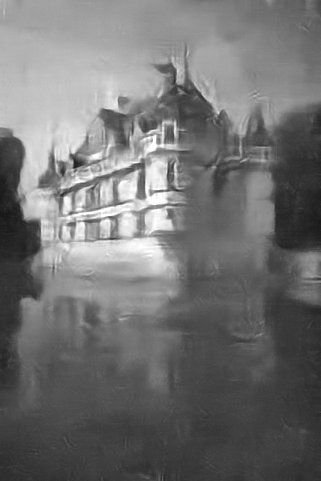}
}
\caption{Poisson noise denoising results comparison. The results are reported by PSNR/SSIM index. Best results are marked.}
\label{peak2}
\end{figure*}

\begin{figure*}[htbp]
\centering
\subfigure[\scriptsize Clean Image]{
\centering
\includegraphics[width=0.17\textwidth]{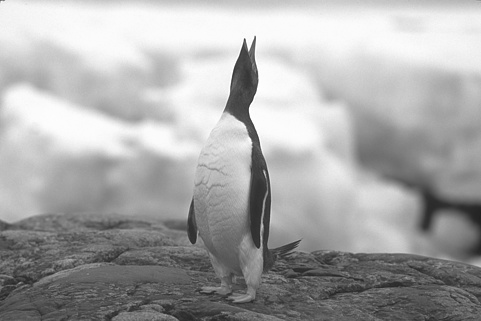}
}
\subfigure[\scriptsize Noisy image. Peak=4]{
\centering
\includegraphics[width=0.17\textwidth]{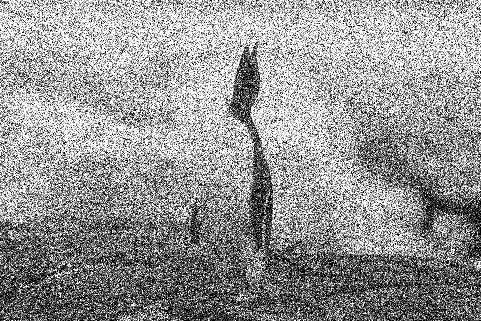}
}
\subfigure[\scriptsize NLSPCA (26.04/0.764)]{
\centering
\includegraphics[width=0.17\textwidth]{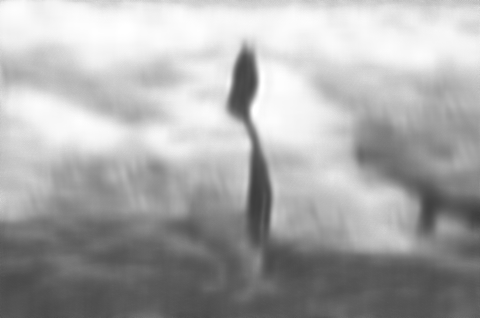}
}
\subfigure[\scriptsize NLSPCAbin (22.94/0.724)]{
\centering
\includegraphics[width=0.17\textwidth]{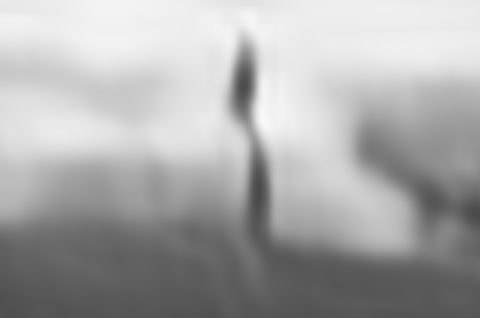}
}
\subfigure[\scriptsize BM3D (26.27/0.684)]{
\centering
\includegraphics[width=0.17\textwidth]{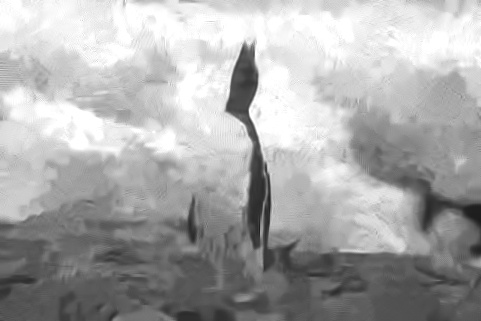}
}\\
\subfigure[\scriptsize BM3Dbin (27.08/0.780)]{
\centering
\includegraphics[width=0.17\textwidth]{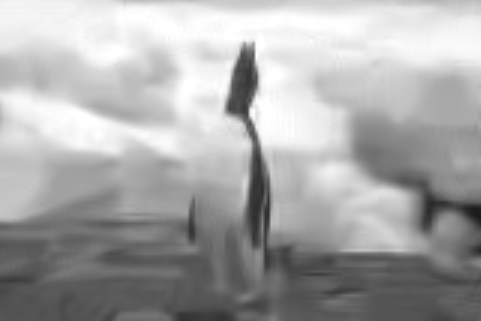}
}
\subfigure[\scriptsize $\mathrm{TNRD}_{5 \times 5}^5$ (27.01/0.738)]{
\centering
\includegraphics[width=0.17\textwidth]{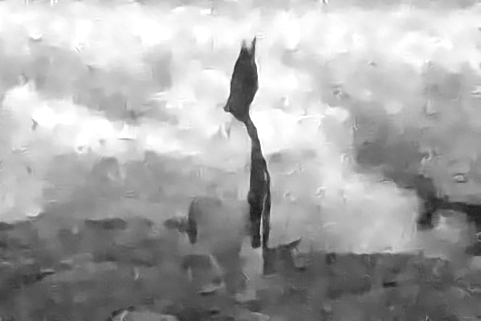}
}
\subfigure[\scriptsize $\mathrm{TNRD}_{7 \times 7}^5$ (27.46/0.762)]{
\centering
\includegraphics[width=0.17\textwidth]{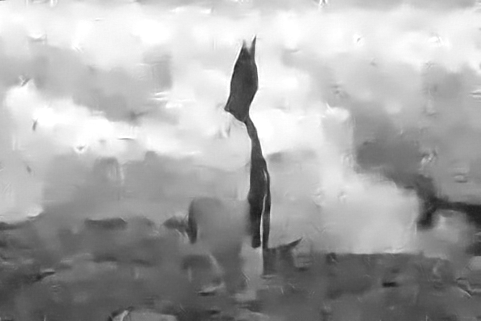}
}
\subfigure[\scriptsize $\mathrm{MSND}_{5 \times 5}^5$ (27.72/0.774)]{
\centering
\includegraphics[width=0.17\textwidth]{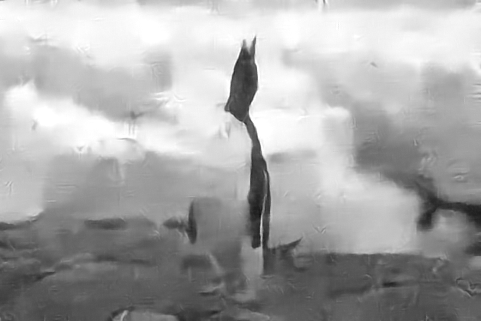}
}
\subfigure[\scriptsize $\mathrm{MSND}_{7 \times 7}^5$ (\textbf{27.85}/\textbf{0.784})]{
\centering
\includegraphics[width=0.17\textwidth]{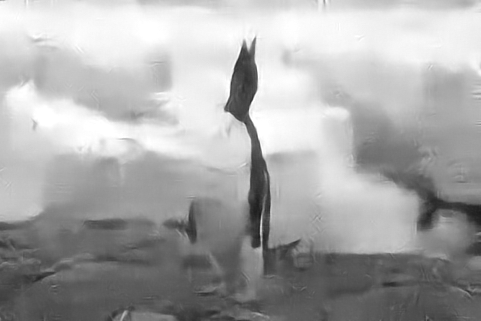}
}
\caption{Poisson noise denoising results comparison. The results are reported by PSNR/SSIM index. Best results are marked.}
\label{peak4}
\end{figure*}

Overall speaking, the denoising performance for Poisson noise denoising is consistent with the Gaussian case. It is worthy noting that the BM3D-based method without binning brings in the typical structured artifacts especially in smooth areas. However, the binning technique yields noisy images with lower noise level, thereby the denoised results in this case is less disturbed by the structured signal-like artifacts. But the adoption of the binning technique leads to resolution reduction which will weaken or even eliminate many image details. In the visual quality, the typical structured artifacts encountered with the single-scale local model do not appear when the proposed method MSND is used, which can noticeably be visually perceived by the comparison in Fig.~\ref{peak1}-Fig.~\ref{peak4}.

Overall speaking, the performance of MSND in terms of PSNR/SSIM is better than the other methods, as shown in Table~\ref{resultshow2}. This indicates that for most images our method is more powerful in the recover quality and geometry feature preservation.

\begin{figure*}[t!]
\centering
\subfigure[peak=1, the filter size is $5\times 5$]{
\centering
\includegraphics[width=0.4\textwidth]{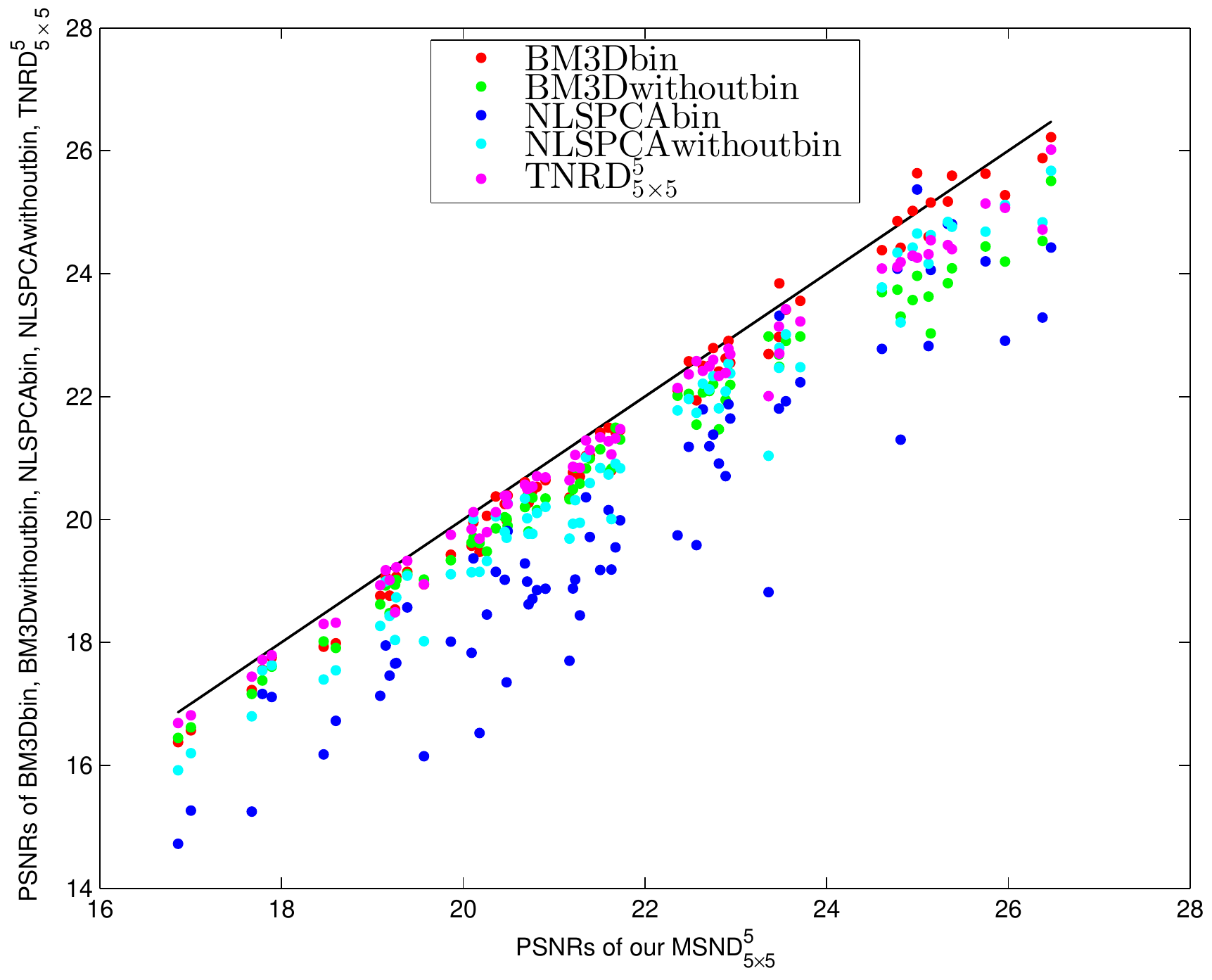}
}
\subfigure[peak=2, the filter size is $5\times 5$]{
\centering
\includegraphics[width=0.4\textwidth]{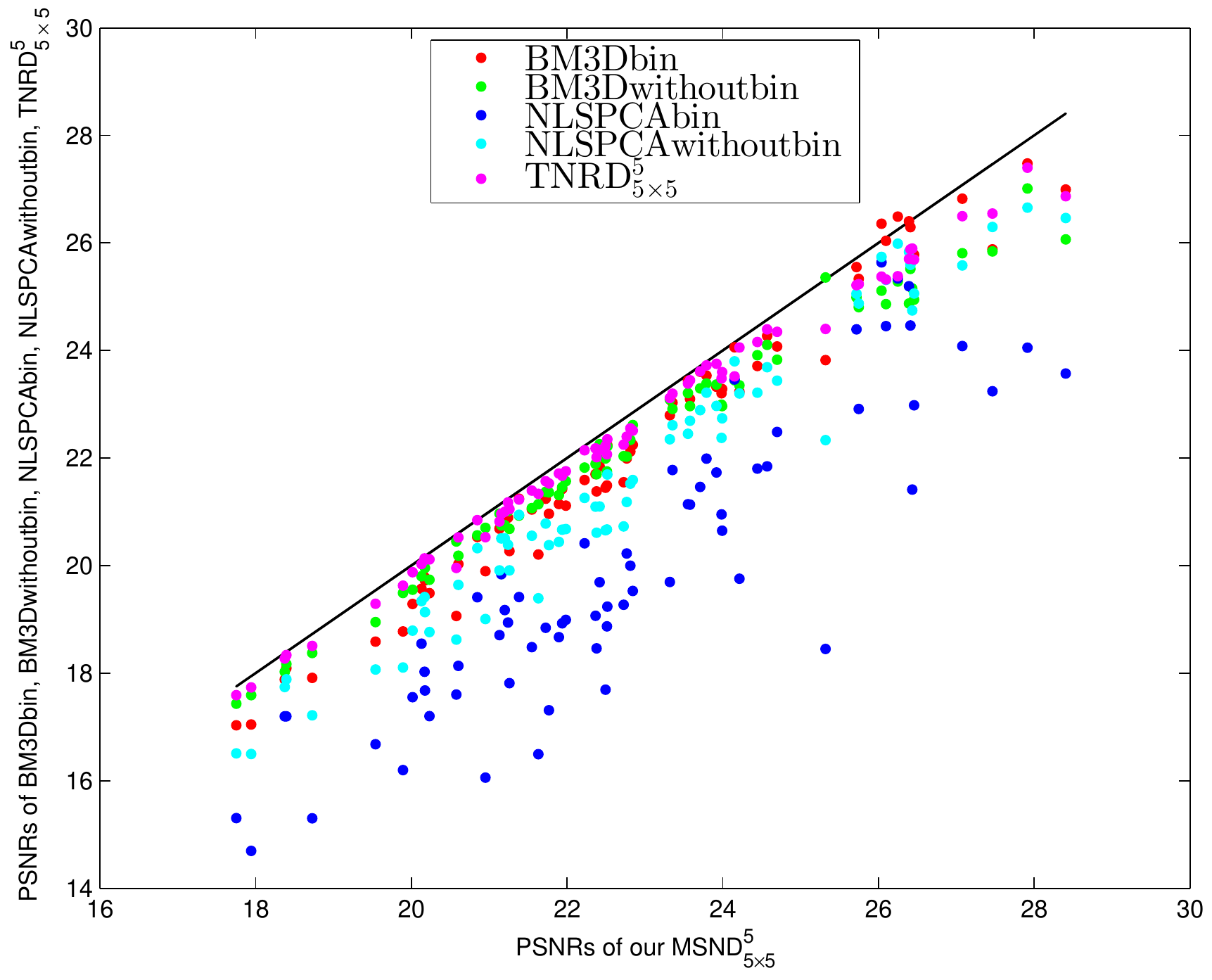}
}\\
\subfigure[peak=1, the filter size is $7\times 7$]{
\centering
\includegraphics[width=0.4\textwidth]{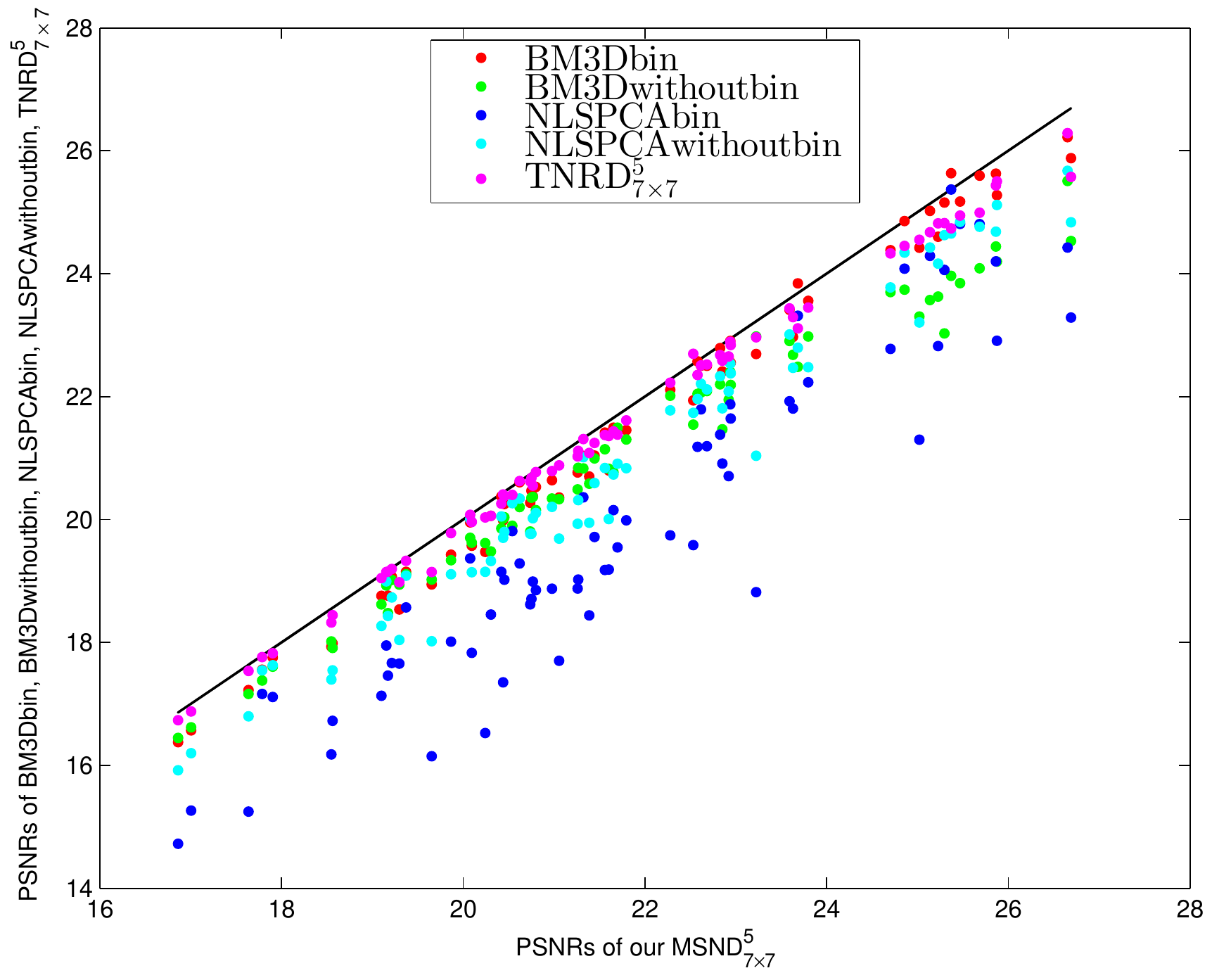}
}
\subfigure[peak=2, the filter size is $7\times 7$]{
\centering
\includegraphics[width=0.4\textwidth]{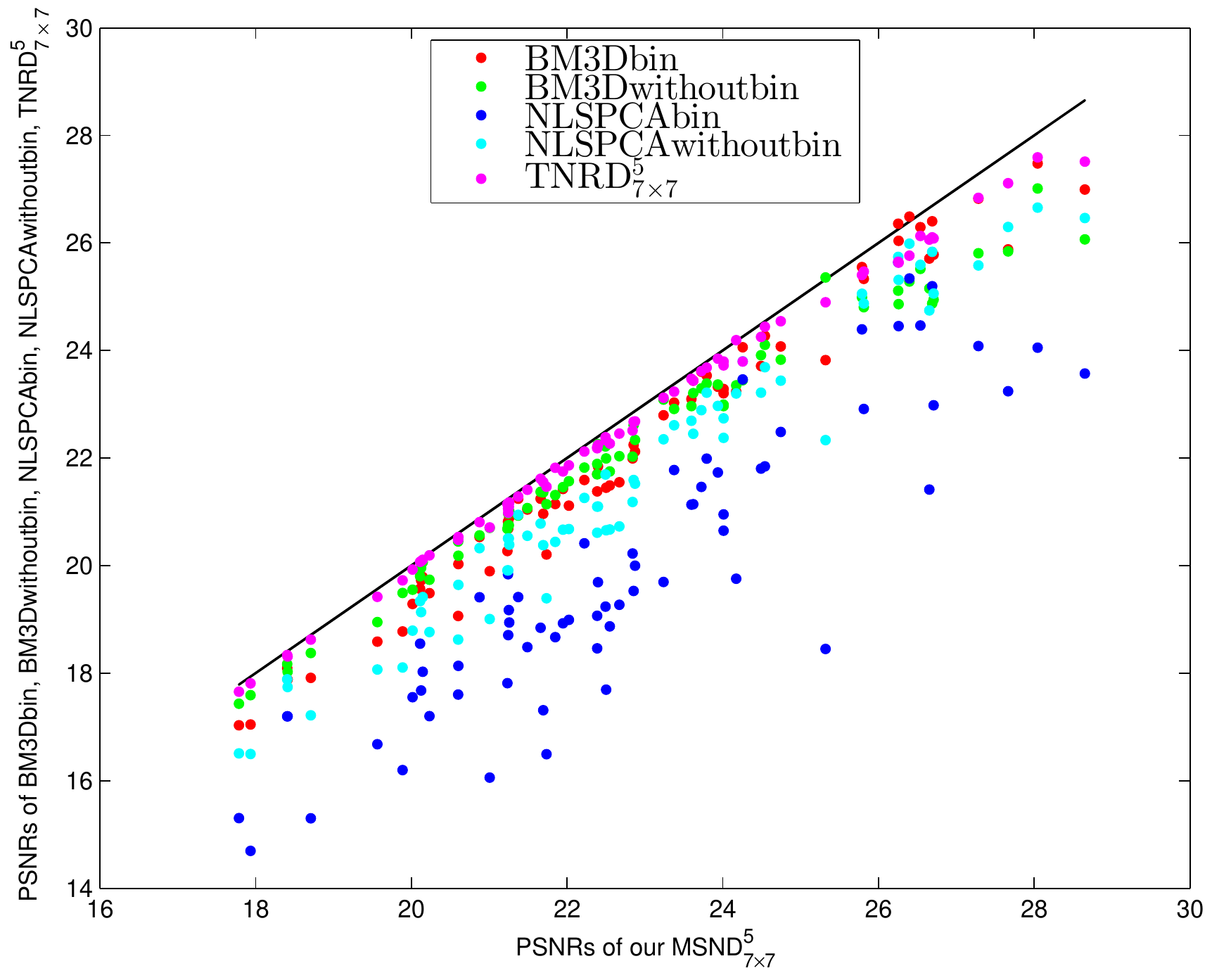}
}
\caption{Scatter plot of the PSNRs over 68 test images produced by the proposed MSND model, BM3Dbin, BM3Dwithoutbin, NLSPCAbin, NLSPCAwithoutbin and TNRD. A point above the diagonal
line (i.e., $y = x$) means a better performance than the proposed model, and a
point below this line indicates a inferior result. One can see that the proposed MSND model outperforms the other methods on almost all test images.}
\label{plotcompare2}
\vspace*{-0.5cm}
\end{figure*}

Figure~\ref{plotcompare2} presents a detailed comparison between our learned MSND model and four state-of-the-art methods over 68 natural
images for noise level peak = 1 and peak = 2, where the diagonal
line $y = x$ means an equal performance to ours. One can see that our method performs better for most images.
\begin{table}[htbp]
\footnotesize
\centering
\begin{tabular}{|c |c |c |c |}
\hline
Method & Peak=1 &Peak=2 &Peak=4\\
\hline
\hline
NLSPCA & 20.90/0.491 &21.60/0.517&22.09/0.535\\
\hline
NLSPCAbin & 19.89/0.466 &19.95/0.467&19.95/0.467\\
\hline
BM3D & 21.01/0.504 &22.21/0.544&23.54/0.604\\
\hline
BM3Dbin & 21.39/0.515 &22.14/0.542&22.87/0.571\\
\hline
$\mathrm{TNRD}_{5 \times 5}^5$& 21.34/0.497 &22.45/0.548&23.58/0.603\\
\hline
$\mathrm{TNRD}_{7 \times 7}^5$& 21.53/0.511 &22.59/0.557&23.80/0.615\\
\hline
$\mathrm{MSND}_{5 \times 5}^5$& 21.71/0.523 &22.78/0.568&23.89/0.620\\
\hline
$\mathrm{MSND}_{7 \times 7}^5$& \textbf{21.76}/\textbf{0.526} &\textbf{22.83}/\textbf{0.573}&\textbf{23.94}/\textbf{0.625}\\
\hline
\end{tabular}
\caption{Comparison of the performance on Poisson denoising of the test algorithms in terms of PSNR and SSIM. Best results are marked.}
\label{resultshow2}
\end{table}

\subsection{Run Time}


\begin{table*}[t!]
\tiny
\begin{center}
\begin{tabular}{|r|c|c|c|c|c|c|c|c|}
\cline{1-9}
& BM3D & WNNM & EPLL&MSEPLL& $\mathrm{TNRD}_{5 \times 5}^5$&$\mathrm{TNRD}_{7 \times 7}^5$&$\mathrm{MSND}_{5 \times 5}^5$&$\mathrm{MSND}_{7 \times 7}^5$\\
\hline\hline
$256 \times 256$ & 1.38 & 154.2&62.9&226.1&   0.63 (\textbf{0.006})&1.48 (\textbf{0.013})&1.62 (\textbf{0.015})&3.76 (\textbf{0.028}) \\
$512 \times 512$ & 4.6 & 612.8 &257.6& 895.7 &1.90 (\textbf{0.018})&4.61 (\textbf{0.043})&4.77 (\textbf{0.043})&11.04 (\textbf{0.101})\\
\cline{1-9}
\end{tabular}
\end{center}
\caption{Typical run time (in second) of the Gaussian denoising methods for images with two different dimensions.
The CPU computation time is evaluated on Intel CPU X5675, 3.07GHz.
The highlighted number is the run time of GPU implementation based on NVIDIA Geforce GTX 780Ti.}
\label{runtime1}
\end{table*}

\begin{table*}[t!]
\scriptsize
\begin{center}
\begin{tabular}{|r|c|c|c|c|c|c|}
\cline{1-7}
& BM3D & NLSPCA & $\mathrm{TNRD}_{5 \times 5}^5$&$\mathrm{TNRD}_{7 \times 7}^5$&$\mathrm{MSND}_{5 \times 5}^5$&$\mathrm{MSND}_{7 \times 7}^5$\\
\hline\hline
$256 \times 256$ & 1.38 & 367.9 &0.65 (\textbf{0.006})&1.52 (\textbf{0.013})&1.63 (\textbf{0.015})&3.80 (\textbf{0.029}) \\
$512 \times 512$ & 4.6 & 1122.1 &1.92 (\textbf{0.019})&4.66 (\textbf{0.044})&4.80 (\textbf{0.043})&11.06 (\textbf{0.101})\\
\cline{1-7}
\end{tabular}
\end{center}
\caption{Typical run time (in second) of the Poisson denoising methods for images with two different dimensions.
The CPU computation time is evaluated on Intel CPU X5675, 3.07GHz.
The highlighted number is the run time of GPU implementation based on NVIDIA Geforce GTX 780Ti.}
\label{runtime2}
\end{table*}

It is worthwhile to note that the proposed model is as efficient as TNRD, and merely contains
convolution of linear filters with an image, which offers high levels of parallelism
making it well suited for GPU implementation.

Table \ref{runtime1} and Table \ref{runtime2} report the typical run time of the proposed model
for the images of two different dimensions for the case of Gaussian denoising and Poisson denoising.
We also present the run time of several competing algorithms\footnote{
All the methods are run in Matlab with single-threaded computation for CPU implementation.
We only consider the version without binning technique.}.

Due to the structural simplicity of the proposed model, it is well-suited to GPU parallel computation. We are able to implement our algorithm on
GPU with ease. It turns out that the GPU implementation
based on NVIDIA Geforce GTX 780Ti can accelerate the inference procedure significantly, as shown in Table \ref{runtime1} and Table \ref{runtime2}. By comparison, we see that the proposed MSND model is generally competitive with the other test methods, especially another multi-scale scheme MSEPLL.

\section{Conclusion}
In this study we employed the multi-scale pyramid image representation to devise a multi-scale nonlinear
diffusion process. The proposed model can efficiently suppress the typical noise artifacts brought by the single-scale local model, especially when the noise level is relatively high. All the parameters in the proposed multi-scale diffusion model, including the filters and the influence functions across scales, are learned from training data through a
loss based approach. Numerical results on Gaussian and Poisson
denoising substantiate that the exploited multi-scale strategy can
successfully boost the performance of the original TNRD model
with single scale. As a consequence, the resulting multi-scale
diffusion models can significantly suppress the typical incorrect
features for those noisy images with heavy noise. Based on standard test dataset, the proposed multi-scale nonlinear diffusion model provides strongly competitive results against state-of-the-art approaches. Moreover, the proposed model bears the properties of simple structure and high efficiency, therefore is well suited to GPU computing. Our future study can be generalization of the proposed multi-scale model into some other image restoration tasks such as deblurring, super-resolution and so on.

{
\small
\bibliographystyle{plain}
\bibliography{references}
}

\end{document}